\documentclass[letterpaper]{article}
\usepackage{auroraworld_preprint}
\usepackage[hyphens]{url}
\usepackage{graphicx}
\usepackage{natbib}
\usepackage{caption}
\usepackage{algorithm}
\usepackage{algorithmic}
\usepackage{booktabs}
\usepackage{tabularx}
\usepackage{array}
\newcolumntype{Y}{>{\raggedright\arraybackslash}X}

\newcommand{\method}{AuroraWorld}
\usepackage{amsmath,amssymb}

\title{\method{}: Structured 4D World Modeling for \\ Exo-to-Ego Video Generation in Robot Manipulation}
\author{
Hao Xue,
Peng-Fei Zhang,
Ying Cheng,\\
Lanyun Zhu,
Lei Zhu,
Heng Tao Shen
}
\affiliations{
{\small\texttt{hao\_xue@foxmail.com, mima.zpf@gmail.com, yingcheng@tongji.edu.cn}}\\
{\small\texttt{lanyun\_zhu@mymail.sutd.edu.sg, leizhu0608@gmail.com, shenhengtao@tongji.edu.cn}}
}

\begin{document}

\maketitle

\begin{abstract}
Egocentric observations provide a robot-centric view of manipulation processes, capturing contact interactions and object-state changes that are important for visuomotor policy learning but difficult to observe from external cameras. However, collecting such data at scale remains expensive due to additional hardware requirements and platform-specific deployment costs. Synthesizing egocentric observations from large-scale exocentric manipulation videos provides a scalable alternative. However, existing methods struggle to synthesize physically consistent and policy-relevant egocentric videos under substantial viewpoint transformations, severe occlusions, complex object interactions, and inherently sparse observations from available exocentric viewpoints.

To address these challenges, we propose \method{}, a 4D world model that infers a structured 4D embodied world state from sparse exocentric observations to enable unobserved egocentric video generation. \method{} characterizes the underlying embodied environment through complementary spatial, temporal, and motion components, capturing scene structure, temporal evolution, and motion dynamics from 3D scene geometry, temporal keyframes, and motion trajectories, respectively. These components provide structured world priors for diffusion generation, enabling geometrically consistent, temporally coherent, and dynamically plausible egocentric videos. We evaluate the proposed method on visual generation and downstream embodied policy learning tasks in both simulated and real robotic environments, demonstrating consistent improvements over existing methods across multiple benchmarks.
\end{abstract}

\section{Introduction}
Embodied intelligence seeks to develop intelligent agents that can perceive, reason about, and interact with the physical world, representing an important direction toward more capable artificial intelligence systems \cite{zhang2025step,jiang2026wsa}. A fundamental challenge in embodied intelligence is learning generalizable robot policies that can transform sensory observations into reliable actions under diverse and complex physical interactions. Recent advances in large-scale robot learning, particularly vision-language-action models \cite{brohan2023rt2,kim2024openvla,khazatsky2025droid,zhang2026dreamvla,zhong2026acot}, have demonstrated the potential of data-driven approaches for acquiring increasingly generalizable manipulation skills. 

\begin{figure}[t]
\centering
\includegraphics[
    width=0.47\textwidth,
]{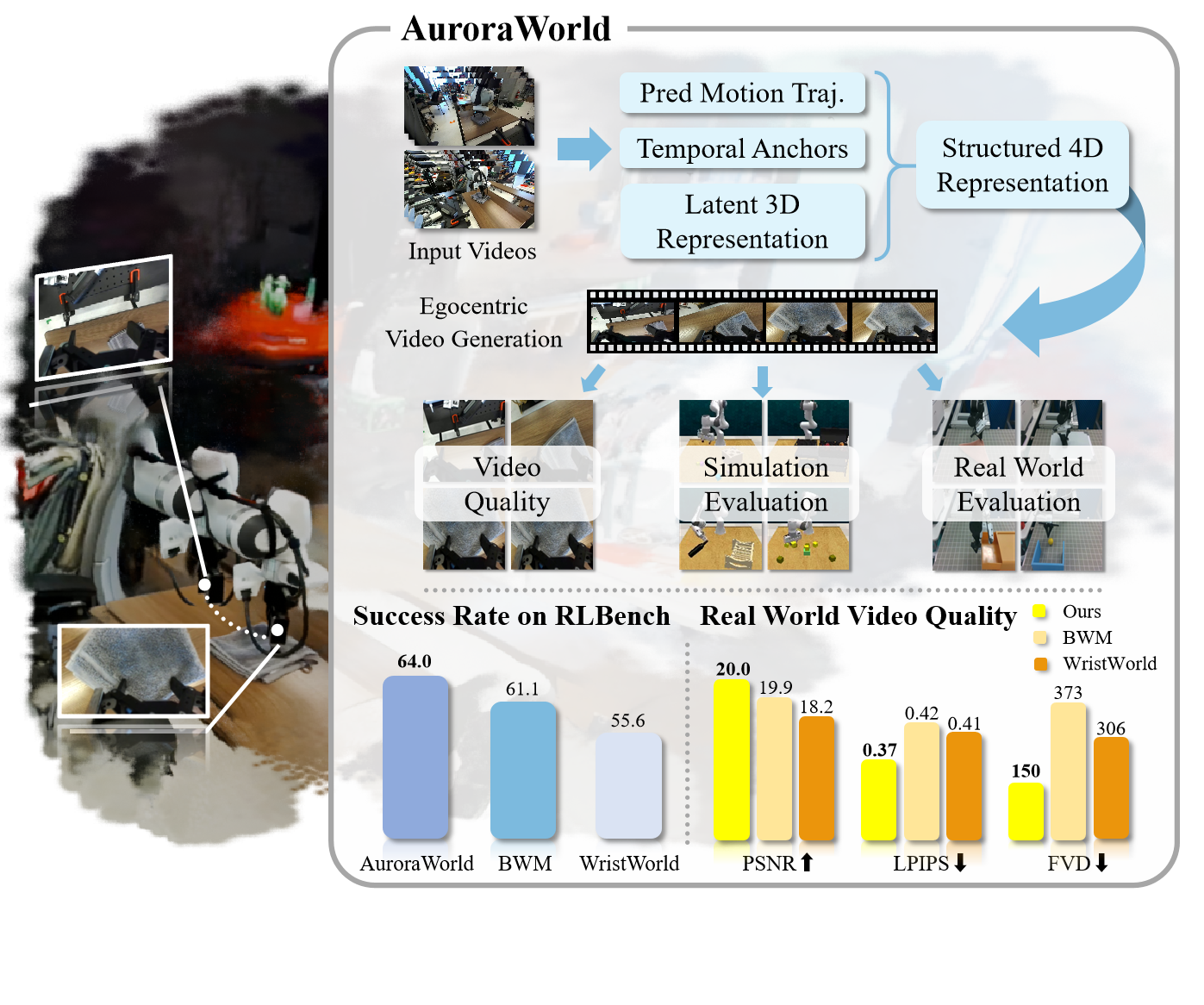}
\caption{\method{} proposes to models the underlying spatial, temporal and motion dynamics, enabling physically plausible exo-to-ego video generation.}
\label{fig:pipeline}
\end{figure}

However, scaling embodied policy learning remains limited by the availability of diverse and high-quality robot interaction data, which is costly and time-consuming to collect in real-world environments. A critical aspect of this data bottleneck is the limited availability of egocentric observations in robot manipulation, such as those captured by wrist-mounted cameras. These observations provide a robot-centric view of manipulation and reveal fine-grained cues, including gripper-object contact and object states, that are valuable for precise control \cite{grauman2022ego4d,khazatsky2025droid}. However, egocentric videos remain less consistently available across robot platforms and tasks, as they require additional sensing setups and dedicated data collection efforts \cite{qian2025wristworld,liu2024exo2ego}. This limited availability of robot-centric observations can hinder the learning of robust and generalizable embodied policies. 

To alleviate this data bottleneck, researchers have explored synthesizing egocentric videos from exocentric manipulation ones. First, some works perform frame-based view synthesis to infer egocentric observations from exocentric views \cite{mildenhall2020nerf,wang2024dust3r,leroy2024mastr,wang2025vggt,szymanowicz2026lagernvs,park2026egoworld,hu2025cascaded}. However, these approaches mainly synthesize individual observations, without capturing temporal evolution and robot-object interaction dynamics. Consequently, they are insufficient for generating temporally coherent manipulation videos involving evolving object states and interactions. Second, recent advances in video generation models enable the synthesis of temporally coherent videos from exocentric observations \cite{liu2024exo2ego,xu2025egoexogen,mahdi2025exo2egosyn,kang2025egox}. Nevertheless, these approaches primarily focus on visual appearance generation, while lacking explicit modeling of the underlying embodied world state required for physically grounded manipulation. Recently, emerging robotic world models \cite{qian2025wristworld} aim to model latent embodied states and the underlying dynamics of physical environments for predicting robot-centric observations. However, most video generation approaches and world models often require additional observational constraints, such as an initial egocentric frame or rich exocentric observations, and do not explicitly model the underlying embodied dynamics \cite{xu2025egoexogen,mahdi2025exo2egosyn,liu2024exo2ego,qian2025wristworld}, to resolve viewpoint ambiguity. However, under sparse exocentric observations, they struggle with large viewpoint changes, severe occlusions, and complex robot-object interactions, making it difficult to generate physically plausible egocentric observations.

To address this challenge, we formulate exo-to-ego synthesis as structured inference of the hidden 4D embodied world representation from sparse exocentric observations. To this end, we propose \method{}, a 4D world model that infers and models the evolving embodied environment through complementary spatial, temporal, and motion components (Figure \ref{fig:pipeline}). Specifically, \method{} captures spatial structures through 3D geometric scene representations, temporal evolution through motion-stabilized keyframes, and interaction dynamics through motion trajectories. These components characterize different aspects of the underlying embodied world and serve as structured world priors for egocentric video generation. By conditioning generation on the inferred 4D representation rather than appearance-level correlations alone, \method{} enables physically consistent and temporally coherent egocentric video synthesis. We evaluate \method{} on visual generation and downstream embodied policy learning tasks in simulated and real robotic environments, demonstrating improved generation quality and policy performance under sparse exocentric observations.

\section{Related Work}
\noindent\textbf{Cross-view Generation.} Cross-view generation aims to synthesize observations from unseen viewpoints given available visual inputs. For static scenarios, recent approaches model scene structures from visual inputs to enable consistent generation from unseen viewpoints, including 3D reconstruction \cite{wang2024dust3r,leroy2024mastr,wang2025vggt,liu2025vggtx,huang2026nas3r,szymanowicz2026lagernvs}, and diffusion-based synthesis \cite{park2026egoworld}. Extending to dynamic scenarios, recent exo-to-ego approaches \cite{liu2024exo2ego,xu2025egoexogen,mahdi2025exo2egosyn,kang2025egox,song2025worldwander,mahdi2026synchrony,luo2024put} leverage conditional video generation models with multi-view, pose-aware, and other forms of conditioning. However, these approaches are primarily developed for general visual scenarios, and perceptual realism is the dominant objective. In contrast, robotic manipulation requires physically grounded egocentric videos that preserve fine-grained geometric relationships, contact interactions, and object-state transitions essential for downstream policy learning. Qian \textit{et al.} \cite{qian2025wristworld} represent an initial effort toward egocentric video synthesis by exploring geometric information. However, their method lacks explicit modeling of key manipulation states or motion dynamics, limiting its applicability under sparse exocentric observations.

\noindent\textbf{Video Generation and Embodied World Models.}
Recent advances in large-scale video generation have achieved significant progress in synthesizing temporally coherent and realistic videos. Latent diffusion models and Transformer-based architectures have become widely adopted paradigms \cite{blattmann2023svd,wan2025wan,kong2024hunyuanvideo,hacohen2025ltxvideo,li2026skyreelsv3,yang2026omnivideo2}. To improve controllability, conditional video generation methods incorporate auxiliary visual signals, such as reference images, depth, camera poses, and geometric structures, to enhance consistency across generated sequences \cite{lee2024videoinpainter,bian2025videopainter,cai2025videocanvas,chen2026depthdirector,cao2026freeorbit4d,hu2026svor}. Building upon the advances, video-generation-based embodied world models are proposed to learn latent environment dynamics and predict future states conditioned on actions \cite{hu2025vpp,zhen2025tesseract,jang2025dreamgen,dharmarajan2025dream2flow,ye2026robodream}. While conventional video generation models primarily optimize visual plausibility and temporal consistency, embodied world models seek to capture latent physical states and action-conditioned dynamics. Recent studies explore strategies, including geometric modeling \cite{zhen2025tesseract,dharmarajan2025dream2flow}, memory mechanisms \cite{guo2025ctrl}, and action conditioning \cite{zhu2025irasim} to improve physical plausibility, consistency, and long-horizon prediction. Embodied world models have been explored as simulators for generating interactive training environments \cite{agarwal2026cosmos,guo2025ctrl}, data engines for scalable synthetic video generation \cite{jang2025dreamgen,ye2026robodream}, and planners for decision making \cite{agarwal2026cosmos,bi2026motus}.

\noindent\textbf{Embodied Policy Learning and Vision-Language-Action Models.}
Embodied policy learning aims to learn generalizable policies that enable robots to perform diverse tasks from sensory observations \cite{chi2025diffusion,kim2026cosmos,khazatsky2025droid,yin2026mivla}. Vision-language-action models have emerged as a promising paradigm by jointly modeling visual observations, language instructions, and robot actions, enabling scalable learning of manipulation policies from heterogeneous datasets \cite{brohan2023rt2,kim2024openvla,khazatsky2025droid,yin2026mivla,zhang2026dreamvla,zhong2026acot}. However, the performance and generalization of these models remain strongly dependent on the availability and diversity of robot interaction data, particularly egocentric observations that capture contact-rich manipulation processes. Collecting such videos at scale remains challenging, motivating approaches that can synthesize policy-relevant egocentric videos from more readily available exocentric observations.

\section{Method}
\subsection{Overview}
\method{} aims to infer a structured 4D embodied world representation from sparse exocentric observations for egocentric video generation. Specifically, given exocentric videos 
$\mathbf{O}^{exo}=\{x^{(i)}_{1:T}\}_{i=1}^{N_{cam}}$
and camera configurations $\mathbf{C}^{exo}$,
\method{} constructs a 4D world representation:
\begin{equation}
\mathbf{W}^{4D}
=
\Phi
(
\mathbf{O}^{exo},
\mathbf{C}^{exo}
),
\end{equation}
where $\Phi(\cdot)$ denotes the 4D representation construction process, and \(\mathbf{W}^{4D}=\{\mathbf{W}^{\mathrm{tp}},\mathbf{W}^{\mathrm{sp}},\mathbf{W}^{\mathrm{mo}}\}\) consists of key temporal information ($\mathbf{W}^{\mathrm{tp}}$), spatial structures ($\mathbf{W}^{\mathrm{sp}}$), and motion dynamics ($\mathbf{W}^{\mathrm{mo}}$), respectively.

The inferred 4D representation, together with exocentric observations, guides a latent diffusion model for egocentric video generation:
\begin{equation}
x^{ego}_{1:T}
\sim
p_{\theta}
(
x^{ego}_{1:T}
|
\mathbf{W}^{4D},
\mathbf{O}^{exo},
\mathbf{C}^{exo}
),
\end{equation}
where $p_{\theta}$ denotes the conditional video generation distribution parameterized by $\theta$.

\begin{figure*}[t]
\centering
\includegraphics[
    width=1\textwidth,
]{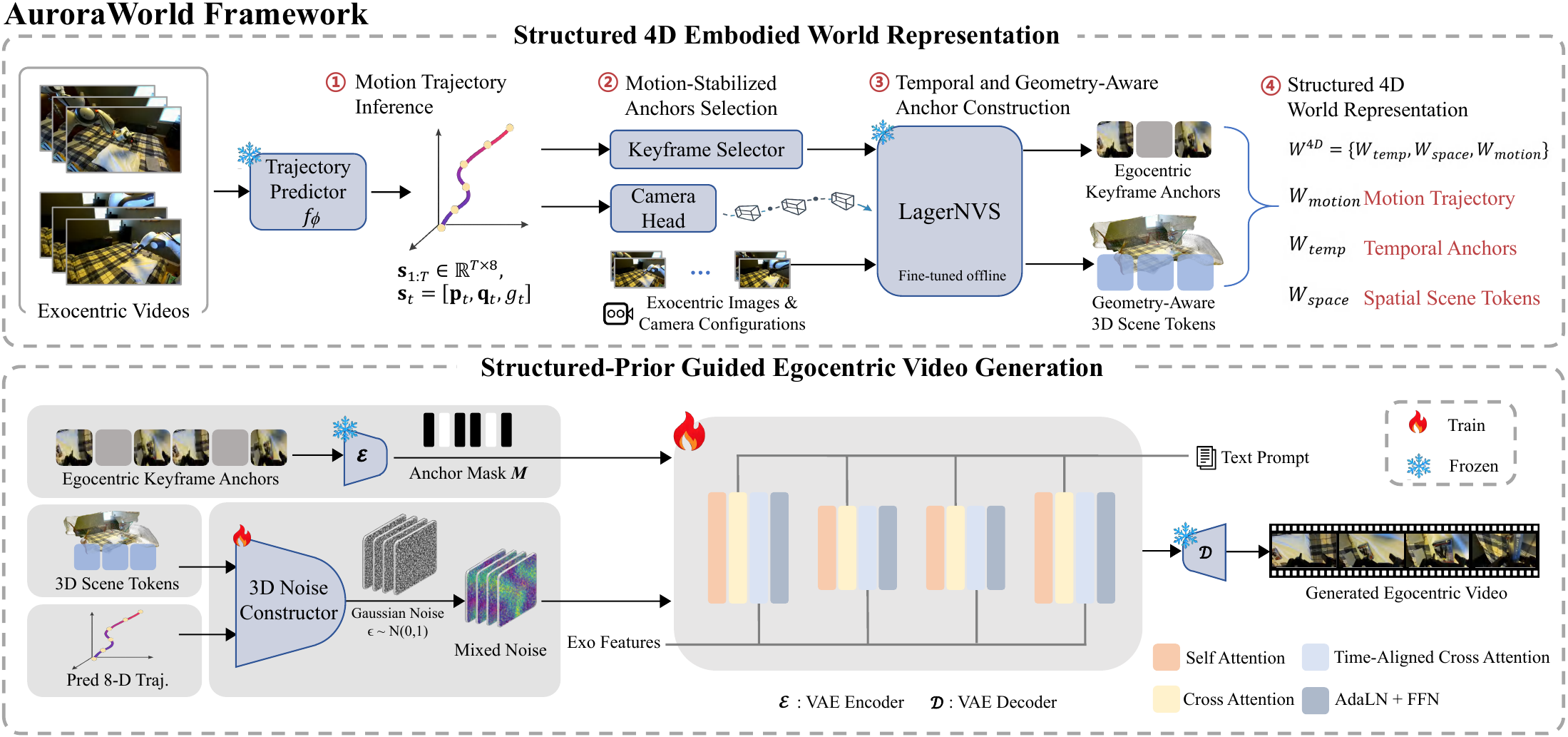}
\caption{
An overview of \method{}. The proposed framework consists of two stages. In the first stage, \method{} infers a structured 4D world representation from sparse exocentric observations. Specifically, it identifies informative temporal keyframes, captures spatial structures, and models motion dynamics from exocentric views. The selected keyframes are then transferred to the egocentric viewpoint and, together with spatial and motion representations, serve as structural priors for video generation. In the second stage, these priors condition the diffusion model to synthesize coherent and physically consistent egocentric videos.}
\label{fig:overview}
\end{figure*}

\subsection{Structured 4D Embodied World Representation}
\subsubsection{Motion Trajectory Inference.}
Robot-object interactions exhibit continuous temporal evolution that characterizes the underlying motion state of embodied environments. In many practical settings, exocentric videos lack recorded ground-truth end-effector trajectories, motivating inference from visual observations. Accordingly, \method{} infers end-effector trajectories from external-view observations, establishing motion anchors for guiding egocentric video generation. Specifically, we adapt an inverse dynamics model (IDM)~\cite{jang2025dreamgen} for the exocentric observation setting to infer the 8-D end-effector trajectory:
\begin{equation}\label{eq2-1}
\mathbf{s}_{1:T}=f_{\phi}(\mathbf{O}^{exo}),
% &\mathbf{s}_t=[\mathbf{p}_t,\mathbf{r}_t,g_t]\in\mathbb{R}^{7},
\end{equation}
where $f_{\phi}$ denotes the adapted inverse dynamics model parameterized by $\phi$. Each trajectory state is represented as:
\begin{equation}\label{eq2-1-1}
\mathbf{s}_t=[\mathbf{p}_t,\mathbf{q}_t,g_t]\in\mathbb{R}^{8},
\end{equation}
where $\mathbf{p}_{t}\in\mathbb{R}^{3}$ denotes the position, $\mathbf{q}_{t}\in\mathbb{R}^{4}$ denotes the unit quaternion, and $g_t$ denotes the gripper state. The inferred end-effector trajectory serves as a motion anchor $\mathbf{W}^{\mathrm{mo}}$ that captures the temporal evolution of the embodied manipulation process. It also guides temporal anchor construction and provides motion guidance for constructing the action- and view-aware 3D noise prior. 

% Adaptation and inference details of the IDM are provided in the appendix.

\subsubsection{Motion-Stabilized Temporal Anchor Construction.}
Temporal anchors capture critical manipulation states while preserving the embodied environment's evolution during egocentric video generation. Rather than densely synthesizing all intermediate viewpoints, introducing redundant constraints and potentially accumulating temporal inconsistencies over long horizons, \method{} selects sparse, informative temporal anchors based on motion transitions and post-transition stabilization. Specifically, the temporal anchor score considers two complementary cues: (1) view transition, measuring whether a candidate frame captures a meaningful change in the end-effector configuration, and (2) motion stabilization, ensuring the selected frame represents a settled manipulation state rather than an intermediate motion phase.

Specifically, we define a transition score $\eta_t$ and a stabilization score $\rho_t$, which are combined to obtain the temporal anchor score:
\begin{equation}\label{eq1-2}
F_t=\mathcal{R}(\eta_t\rho_t),
\end{equation}
where $\mathcal{R}(\cdot)$ denotes robust per-clip normalization to $[0,1]$. 

The transition score $\eta_t$ measures the accumulated end-effector pose change over a preceding temporal interval $\Delta$, indicating whether the candidate frame represents a new and informative viewpoint:
\begin{equation}\label{eq1-3}
\eta_t=
\mathcal{R}
\left(
d_p(\mathbf{p}_{t},\mathbf{p}_{t-\Delta})
+
\lambda_q d_q(\mathbf{q}_{t-\Delta},\mathbf{q}_{t})
\right),
\end{equation}
where $d_p(\cdot)$ denotes Euclidean distance, $d_q(\cdot)$ denotes geodesic rotation distance, and $\lambda_q$ balances translation and rotation changes.

The stabilization score $\rho_t$ evaluates whether the end-effector has reached a stable configuration after the transition. To measure it, we first compute the one-step translation and rotation changes:
\begin{equation}\label{eq1-4}
\begin{split}
\delta^{p}_{t} &= d_p(\mathbf{p}_{t},\mathbf{p}_{t-1}),\\
\delta^{r}_{t} &=
d_q( \mathbf{q}_{t},\mathbf{q}_{t-1}).
\end{split}
\end{equation}

To reduce frame-level motion fluctuations, we apply temporal smoothing to the translation and rotation changes, obtaining the smoothed changes $\bar{\delta}^{p}_{t}$ and $\bar{\delta}^{r}_{t}$. They are then aggregated to obtain the instantaneous motion signal:
\begin{equation}\label{eq1-5}
o_t=\frac{1}{2}\left[\mathcal{R}(\bar{\delta}^{p}_{t})+
\mathcal{R}(\bar{\delta}^{r}_{t})\right].
\end{equation}

The stabilization score is then computed over a short future temporal window $\mathcal{N}_t$:
\begin{equation}\label{eq1-6}
\begin{split}
\rho_t = 1-\mathcal{R}\!\left(\frac{1}{|\mathcal{N}_t|}\sum_{u\in\mathcal{N}_t}o_u
\right),
\end{split}
\end{equation}
where higher values indicate that the end-effector has reached a locally stable configuration after the transition. 

Based on the temporal anchor scores $F_t$, we select top-$k$ interior frames and always include the first and last frames, yielding an anchor index set $\mathcal{K}$ of $k$+2 frames. These locations identify critical states of the evolving embodied environment for constructing the structured 4D embodied representation. Since exocentric observations lack the egocentric visual information required for embodied observation generation, \method{} adapts LagerNVS~\cite{szymanowicz2026lagernvs} to transform the corresponding exocentric observations into egocentric keyframes. For each selected timestep $t\in\mathcal{K}$, a lightweight state-conditioned camera head first predicts the target ego-view camera pose from the inferred end-effector state. The adapted LagerNVS then synthesizes the corresponding egocentric keyframe conditioned on the exocentric observation, its camera configuration, and the predicted target-view pose:
\begin{equation}
\begin{aligned}
\hat{\mathbf{C}}^{ego}_{t}
&=H_{\mathrm{cam}}(\mathbf{s}_{t}),\\
\hat{x}^{ego}_{t}
&=\mathcal{G}_{\mathrm{LagerNVS}}
\left(
\mathbf{O}^{exo}_{t},
\mathbf{C}^{exo}_{t},
\hat{\mathbf{C}}^{ego}_{t}
\right).
\end{aligned}
\end{equation}
Here, $H_{\mathrm{cam}}$ denotes the state-conditioned camera head, and $\mathcal{G}_{\mathrm{LagerNVS}}$ denotes the adapted LagerNVS model. The synthesized keyframes constitute
$\mathbf{W}^{\mathrm{tp}}=\{\hat{x}^{ego}_{t}\mid t\in\mathcal{K}\}$.

\subsubsection{Geometry-Aware Spatial Anchor Construction}
Large exo-to-ego viewpoint transformations introduce substantial spatial ambiguity, making geometry-consistent scene understanding essential for generating physically plausible egocentric videos. To construct the spatial component of the structured 4D representation, \method{} extracts latent geometry-aware scene representations from sparse exocentric observations. Specifically, we leverage LagerNVS~\cite{szymanowicz2026lagernvs} equipped with camera tokens to obtain latent 3D scene tokens from sparse exocentric views. These latent representations encode viewpoint-consistent scene structures and preserve geometric relationships among objects and the environment without requiring explicit 3D reconstruction. The resulting 3D tokens constitute the spatial component $\mathbf{W}^{\mathrm{sp}}$, enabling synthesized egocentric videos to maintain consistent scene structures under large viewpoint changes.

\subsection{Structured-Prior Guided Egocentric Video Generation}
Given the inferred structured 4D embodied representation $\mathbf{W}^{4D}$, \method{} uses it as a structured prior to guide egocentric video synthesis through a latent diffusion-based video generation model. Specifically, the temporal, spatial, and motion components provide complementary constraints for video generation through temporal observation conditioning and spatial-motion-aware latent initialization during the denoising process.

\subsubsection{Temporal Anchor State Conditioning.}
Given constructed keyframes $\mathbf{W}^{\mathrm{tp}}=
\{\hat{x}^{ego}_{t}\mid t\in\mathcal{K}\}$, we place each anchor at its original temporal location and construct a sparse conditioning video:
\begin{equation}
\label{eq:anchor_video_condition}
\mathbf{a}_{t}=
\begin{cases}
\hat{x}^{ego}_{t}, & t\in\mathcal{K},\\
\mathbf{0}, & t\notin\mathcal{K}.
\end{cases}
\end{equation}

Meanwhile, we introduce a binary anchor mask to indicate valid temporal locations:
\begin{equation}
\label{eq:anchor_mask}
\mathbf{M}_{t}=\mathbb{I}[t\in\mathcal{K}],
\end{equation}
where $\mathbf{M}_{t}=1$ indicates the presence of an egocentric temporal anchor at the corresponding temporal location, while $\mathbf{M}_{t}=0$ denotes an unconditioned slot.

The sparse temporal observation condition is encoded using a VAE encoder and concatenated with the aligned mask:
\begin{equation}
\label{eq:encoded_anchor_condition}
\mathbf{y}=
\operatorname{Concat}_{\mathrm{ch}}
\left(\mathcal{D}(\mathbf{M}_{1:T}),E_{\mathrm{VAE}}(\mathbf{a}_{1:T})\right),
\end{equation}
where \(E_{\mathrm{VAE}}\) denotes a latent VAE encoder,
\(\operatorname{Concat}_{\mathrm{ch}}\) represents  channel-wise concatenation,
and \(\mathcal{D}\) broadcasts and temporally packs the frame-level mask to match the VAE latent layout.

\subsubsection{Spatial-Motion Anchor-Guided Latent Initialization}
Current generative approaches for embodied world modeling typically initialize the generation process from unstructured Gaussian noise, where the initial state contains no explicit
information about the underlying environment or embodied interaction. This would lead to ambiguous and physically inconsistent generation under large viewpoint changes and sparse observations. To address this limitation, \method{} constructs a spatial-motion guided latent prior from the inferred spatial and motion components $\mathbf{W}^{\mathrm{sp}}$ and $\mathbf{W}^{\mathrm{mo}}$ and injects it into the initial diffusion state.

Specifically, given the latent spatial representation $\mathbf{W}^{\mathrm{sp}}$ extracted from sparse exocentric observations, we first transform it into the spatial feature space:
\begin{equation}\label{eq2-2}
\mathbf{G}
=
\phi_{\mathrm{scene}}
(
\mathbf{W}^{\mathrm{sp}}
).
\end{equation}

Meanwhile, given the inferred component $\mathbf{W}^{\mathrm{mo}}$, we construct a spatial-motion conditioned query for each latent timestep \(\ell\) and spatial location \((i,j)\):
\begin{equation}\label{eq2-3}
\mathbf{u}_{\ell ij}=\mathrm{LN}\left(\phi_{\mathrm{qry}}\left([\phi_{\mathrm{grid}}(i,j);\phi_{\mathrm{act}}(\mathbf{W}^{\mathrm{mo}}_{\ell})]
\right)
\right).
\end{equation}
The query encodes the desired observation conditioned on spatial location and embodied motion, while the end-effector pose within the motion representation implicitly provides the corresponding ego-view configuration.

The spatial-motion query retrieves relevant geometric information from $\mathbf{G}$ through cross-attention:
\begin{equation}\label{eq2-4}
\mathbf{w}_{\ell ij} = \operatorname{Attn}
( \mathbf{u}_{\ell ij}, \mathbf{G}, \mathbf{G}).
\end{equation}

This operation associates motion evolution with corresponding spatial structures, producing a view- and manipulation-aware latent prior:
% The retrieved feature is projected into the VAE latent space:
\begin{equation}\label{eq2-5}
\begin{split}
\boldsymbol{\epsilon}^{sm}_{\ell ij} = \phi_{\mathrm{out}}(\mathbf{w}_{\ell ij}) \in \mathbb{R}^{C_z},
\end{split}
\end{equation}
where $\phi_{\mathrm{out}}$ projects each retrieved feature
into the VAE latent channels. Stacking
$\boldsymbol{\epsilon}^{sm}_{\ell ij}$ over all latent positions and
standardizing the resulting tensor per sample over all latent
elements gives
$\bar{\boldsymbol{\epsilon}}^{sm}\in
\mathbb{R}^{C_z\times T_z\times H_z\times W_z}$.
We define the structured terminal latent as
\begin{equation}\label{eq2-8}
\mathbf{z}^{mix}_{1}
=
\sqrt{1-\lambda^2}\,\boldsymbol{\epsilon}
+
\lambda\bar{\boldsymbol{\epsilon}}^{sm},
\end{equation}
where $\boldsymbol{\epsilon}$ has i.i.d.\ standard-normal entries with the
same tensor shape and
$\lambda\in[0,1]$ controls the amplitude of the structured
component. The resulting initialization guides diffusion sampling from a structured latent state that incorporates scene geometry and embodied motion, rather than from an unconstrained stochastic latent.

\subsubsection{Training.}
The structured 4D world representation provides the initial state prior and observation constraints for egocentric video generation. Specifically, the spatial and motion components $\mathbf{W}^{\mathrm{sp}}$ and $\mathbf{W}^{\mathrm{mo}}$ are first integrated into a structured terminal latent $\mathbf{z}^{mix}_{1}$, which replaces the conventional Gaussian initialization in the flow-matching process. During training, the intermediate latent at timestep $\tau$ is obtained as:
\begin{equation}
\label{eq2-10}
\mathbf{z}_{\tau}
=
(1-\tau)\mathbf{z}_{0}
+
\tau\mathbf{z}^{mix}_{1},
\end{equation}
where $\mathbf{z}_{0}$ denotes the clean egocentric video latent.

The diffusion model is optimized to predict the corresponding velocity field:
\begin{equation}
\label{eq2-11}
\mathcal{L}=
\mathbb{E}_{\mathbf{z}_{0},\boldsymbol{\epsilon},\tau}\left[\left\|v_{\theta}\left(
\mathbf{z}_{\tau},\mathbf{y},\mathcal{E},\mathbf{p}_{\mathrm{text}},\tau\right)-\left(\mathbf{z}^{mix}_{1}-\mathbf{z}_{0}\right)\right\|_2^2\right],
\end{equation}
where $\mathcal{E}$ denotes the time-aligned exocentric visual feature sequence, and $\mathbf{p}_{\mathrm{text}}$ denotes the text prompt embedding. \(v_{\theta}\) denotes the flow-matching velocity field parameterized by the video generation backbone. $\tau\in[0,1]$ denotes the continuous flow timestep that controls the interpolation between the clean video latent and the structured terminal latent.

\section{Experiments}
\label{sec:experiments}

\subsection{Experimental Setup}
\noindent\textbf{Datasets.} To comprehensively evaluate \method{}  from video generation to embodied policy learning and physical-world validation, we conduct experiments on three datasets. DROID \cite{khazatsky2025droid} is used for large-scale real-world exo-to-ego video generation, where we select 17k successful manipulation episodes with reliable camera calibration and reserve an additional 200 episodes for evaluation.  RLBench \cite{james2019rlbench} is used for downstream policy
learning evaluation under controlled simulation on an eight-task subset of the PerAct suite \cite{shridhar2023peract}. We further collect an
in-house ARX X5 robot dataset for physical-world validation, which contains 600 real-world manipulation episodes across six tabletop tasks.

\noindent\textbf{Baselines.}
Several representative approaches are selected for comparison: (1) frame-wise synthesis methods, including VGGT~\cite{wang2025vggt} and LagerNVS~\cite{szymanowicz2026lagernvs}; (2) general video generation and world modeling approaches, including SVD~\cite{blattmann2023svd} and BWM (Boundless World Model)~\cite{boundless2026}; and (3) exo-to-ego generation methods, including Exo2Ego-V~\cite{liu2024exo2ego} and WristWorld~\cite{qian2025wristworld}. Boundless World Model is a 5B-parameter action-conditioned video generation model. We provide it with the ground-truth first egocentric frame and condition it on the dense trajectory inferred by our IDM from the source exocentric observations, providing a strong baseline to evaluate the effectiveness of our structured 4D world modeling.

\begin{table}[t]
\centering
\small
\setlength{\tabcolsep}{3pt}
\begin{tabular}{lccccc}
\toprule
% & \multicolumn{5}{c}{DROID}\\
% \cmidrule(lr){2-6}
Method 
& PSNR $\uparrow$ & SSIM $\uparrow$ & LPIPS $\downarrow$ & FID $\downarrow$ & FVD $\downarrow$ \\
\midrule
VGGT
& 8.93 & 0.206 & 0.784 & 208.98 & 834.02 \\

LagerNVS
& 15.05 & 0.5130 & 0.4332 & 40.991 & 245.97 \\

SVD
& 12.84 & 0.4971 & 0.5294 & 31.874 & 363.84\\

BWM
& 15.77 & 0.5012 & 0.3309 & 25.632 & 70.073\\

Exo2Ego-V
& 15.43 & 0.4909 & 0.4020 & 22.731 & 130.11 \\

WristWorld
& 13.68 & 0.4204 & 0.4527 & 30.267 & 141.98 \\

\method{}
& \textbf{17.87} & \textbf{0.5800} & \textbf{0.2648} & \textbf{15.437} & \textbf{60.171}\\

\bottomrule
\end{tabular}
\caption{Quantitative comparison of egocentric video generation on DROID. Best results are highlighted in \textbf{bold}.}
\label{tab:visual}
\end{table}

\begin{figure*}[t]
\centering
\includegraphics[
    width=0.96\textwidth,
    % keepaspectratio
]{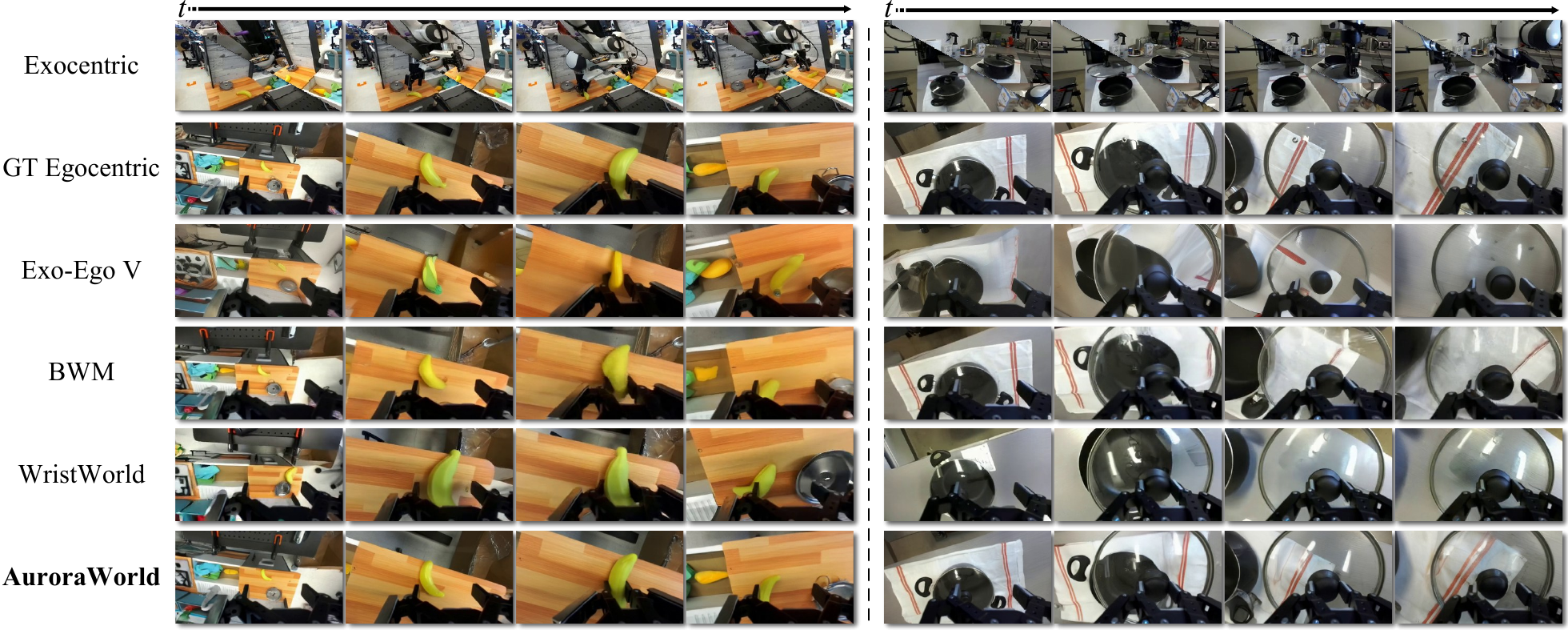}
\caption{Qualitative comparison on DROID. Rows correspond to exocentric left/right ground-truth views, egocentric ground-truth views, and generated egocentric views from \method{} and competing methods, including WristWorld, BWM, and Exo-Ego-V. The results show that \method{} generates egocentric videos with improved geometric consistency, appearance fidelity, and temporal coherence compared with existing approaches.
}
\label{fig:droid_qualitative}
\end{figure*}

\noindent\textbf{Evaluation Metrics.}
For video generation quality, we adopt PSNR, SSIM, LPIPS, FID, and FVD as quantitative metrics. PSNR and SSIM measure pixel-level fidelity and structural similarity between synthesized and ground-truth frames, respectively, while LPIPS evaluates perceptual similarity based on deep visual features. FID measures the distributional distance between generated and real image features, reflecting visual realism, whereas FVD evaluates the distributional discrepancy between generated and real video features to assess temporal dynamics and consistency. For downstream policy learning, we use task success rate as the evaluation metric to assess the effectiveness of synthesized egocentric observations for visuomotor control.

\noindent\textbf{Implementation Details.} We implement \method{} based on the 1.3B Wan-Fun-InP backbone~\cite{wan2025wan}. For parameters, we set $\mathcal{N}_t=8$, $\Delta=3$, $k=3$, $\lambda=0.5$, and $\lambda_q=0.1$.

\begin{table}[t]
\centering
\small
\setlength{\tabcolsep}{10pt}
\begin{tabular}{lccc}
\toprule
Policy
& WristWorld
& BWM
& \method{} \\
\midrule
RVT-2
& 55.60
& 61.10
& \textbf{64.00} \\

$\pi_{0.5}$
& 36.85
& 37.70
& \textbf{40.15} \\
\bottomrule
\end{tabular}
\caption{
RLBench policy success rates (\%) trained with different egocentric observations.}
\label{tab:rlbench_policy}
\end{table}

\begin{table}[t]
\small
\centering
\setlength{\tabcolsep}{3.6pt}
\begin{tabular}{lccccc}
\toprule
Method & PSNR$\uparrow$ & SSIM$\uparrow$ & LPIPS$\downarrow$ & FID$\downarrow$ & FVD$\downarrow$\\
\midrule
WristWorld & 18.24 & 0.654 & 0.414 & 22.71 & 306.34\\
BWM  & 19.88 & \textbf{0.709} & 0.422 & 39.99 & 373.72\\
\method{} & \textbf{20.01} & 0.681 & \textbf{0.375} & \textbf{16.91} & \textbf{149.93}\\
\bottomrule
\end{tabular}
\caption{Quantitative comparison of egocentric video generation on ARX X5. Best results are highlighted in \textbf{bold}.}
\label{tab:wrist}
\end{table}

\subsection{Experimental Results and Analysis}
\noindent\textbf{Video Generation Evaluation.}
We first compare the video generation quality of different methods, with results on DROID reported in Table~\ref{tab:visual}. The results lead to the following observations. First, although VGGT and LagerNVS leverage geometric information for view synthesis, they perform frame-wise generation and lack explicit modeling of temporal dependencies and motion continuity across consecutive frames, leading to inferior video generation performance compared with video-based approaches. Second, the proposed method outperforms existing models on most evaluated metrics by jointly modeling spatial structures, temporal evolution, and motion dynamics, thereby generating more coherent and physically consistent egocentric videos. In addition, although BWM benefits from a larger backbone and uses ground-truth initial egocentric frames as conditioning inputs, the proposed method still achieves better performance. This highlights the advantage of explicitly modeling structured 4D world representations.

\noindent\textbf{Downstream Policy Evaluation in Simulation.}
We further evaluate whether synthesized egocentric videos provide policy-relevant observations for downstream visuomotor learning using RVT-2~\cite{goyal2024rvt2} and ${\pi_{0.5}}$~\cite{black2025pi05} as downstream policy backbones. In the experiments, we consider 8 manipulation tasks on the RLBench benchmark~\cite{james2019rlbench}, including non-prehensile manipulation, pick-and-place, and precision insertion. Policies are trained from expert demonstrations and evaluated over 10 runs, with 25 trials per task in each run. The results are reported in Table~\ref{tab:rlbench_policy}. In the table, {WristWorld}, {BWM}, and {\method{}} denote training with synthesized egocentric videos generated by the corresponding methods. For a controlled comparison, all three synthesized-view settings use the same action labels derived from the robot state trajectories inferred by our IDM from the source exocentric demonstrations; therefore, they differ only in the synthesized egocentric observations used for policy training. From the results, we observe that \method{} achieves the highest average success rate among the synthesized-view methods with both policy backbones. With RVT-2, \method{} achieves 64.00\%, outperforming BWM and WristWorld by 2.90 and 8.40 percentage points, respectively. With \(\pi_{0.5}\), \method{} achieves 40.15\%, exceeding BWM and WristWorld by 2.45 and 3.30 percentage points, respectively. These results demonstrate that, under the same IDM-inferred action supervision, egocentric observations synthesized by \method{} provide more effective training data for downstream visuomotor policies.

\begin{table}[t]
\centering
\small
\setlength{\tabcolsep}{1.2pt}
\renewcommand{\arraystretch}{1.2}
\begin{tabular}{lccccc}
\hline
Task & w/o Ego & WW & BWM & \method{} & Real Ego \\
\hline
Fold Towel  & 2/25 & 11/25& 16/25 &15/25 & 17/25 \\
Pull Drawer & 0/25 & 9/25& 12/25 &15/25 & 15/25 \\
Stack Cups & 0/25 & 0/25 & 1/25 &0/25 & 12/25 \\
Push Ball & 8/25& 8/25 & 9/25 &11/25 & 14/25 \\
Push Cube & 7/25 & 13/25 & 13/25 &15/25 & 18/25 \\
Erase Whiteboard & 7/25 & 7/25 & 9/25 &11/25 & 12/25 \\
\hline
Average & 16.0\% & 32.0\% & 40.0\% &44.7\% & 58.7\% \\
\hline
\end{tabular}
\caption{Real-robot policy evaluation on the ARX X5. WW denotes Wristworld.}
\label{tab:arx_real_robot}
\end{table}

\noindent\textbf{Real-Robot Policy Evaluation.}
To evaluate the real-robot applicability of \method{}, we conduct experiments on the ARX X5 platform, with the results summarized in Tables~\ref{tab:wrist} and~\ref{tab:arx_real_robot}. Table~\ref{tab:wrist} shows that \method{} achieves the best video generation performance. Furthermore, Table~\ref{tab:arx_real_robot} evaluates policies trained under five observation settings: without egocentric observations ({w/o Ego}), with egocentric observations synthesized by WristWorld, BWM, or \method{}, and with real egocentric observations ({Real Ego}). Across all five settings, policy training uses action targets derived from IDM-inferred trajectories. Compared with {w/o Ego}, training with synthesized egocentric observations consistently improves policy performance, indicating that the generated views recover visual information relevant to downstream manipulation. Among the synthesized-view settings, \method{} achieves the highest success rate, outperforming the second-best method by 4.7 percentage points. Together, these results demonstrate that synthesized egocentric observations can effectively recover policy-relevant visual information and improve visuomotor policy learning, while higher-quality synthesis further translates into better downstream policy performance. Nevertheless, \method{} still exhibits a performance gap compared with the ground-truth setting on certain tasks, such as cup stacking. This task requires highly accurate object pose estimation and contact-state understanding, for which synthesized observations may not yet provide sufficient precision for fine-grained spatial alignment. We leave addressing this limitation for future work.

\begin{table}[t]
\centering
% \small
% \footnotesize
\setlength{\tabcolsep}{2pt}
\renewcommand{\arraystretch}{1}
\begin{tabular}{@{}ccccccccr@{}}
\toprule
% \multicolumn{3}{c}{Conditioning}
% & \multicolumn{5}{c}{Generation quality} \\
% \cmidrule(lr){1-3}
% \cmidrule(lr){4-8} temp, space, motion
$\mathbf{W}^{\mathrm{sp}}$
& $\mathbf{W}^{\mathrm{tp}}$
& $\mathbf{W}^{\mathrm{mo}}$
& PSNR $\uparrow$
& SSIM $\uparrow$
& LPIPS $\downarrow$
& FID $\downarrow$
& FVD $\downarrow$ \\
\midrule
--  & -- & --   & 12.96  & 0.43 & 0.50 & 30.78 & 192.93 \\
\checkmark  & -- & --   & 13.09  & 0.50 & 0.49 & 24.69 & 168.45 \\
--         & \checkmark &  --  & 16.69 & 0.55 & 0.31 & 16.95 & 76.12 \\
--   & -- & \checkmark  & 14.52 & 0.49 & 0.42 & 21.99 & 129.28 \\
\checkmark & -- &  \checkmark & 14.38 & 0.49 & 0.42 & 20.47 & 110.43\\
% --         & \checkmark &  --  & 17.45 & 0.57 & 0.28 & 15.79 & 64.66 \\
% \checkmark & \checkmark &  --  & 17.06 & 0.57 & 0.30 & 15.21 & 68.03 \\
-- & \checkmark &  \checkmark  &17.45 &0.57 &0.28 & 15.79 & 64.66 \\
\checkmark & \checkmark &  --  & 17.10 & 0.56 & 0.29 & 15.68 & 65.22 \\
\checkmark & \checkmark &  \checkmark & 17.87 & 0.58 &  0.26 & 15.44 & 60.17 \\
% --         & --         & --         &  GT    & 14.29 & 0.504 & 0.402 & 40.59 & 153.4 \\
% --         & --         & \checkmark &  GT    & 17.73 & 0.585 & 0.273 & 15.86 & 71.45 \\
% \checkmark & --         & \checkmark &  GT    & 18.23 & 0.596 & 0.253 & 14.86 & 63.66 \\
% --         & \checkmark & \checkmark &  GT    &18.98 & 0.613 & 0.226 & 14.49 & 54.31 \\
% \checkmark & \checkmark & \checkmark &  GT    &\textbf{19.17} & \textbf{0.622} & \textbf{0.219} & \textbf{14.16} & \textbf{52.11} \\
\bottomrule
\end{tabular}
\caption{Ablation study of different components.}
\label{tab:ablation_modules}
\end{table}

\begin{table}[t]
\centering
% \small
% \footnotesize 
\setlength{\tabcolsep}{2pt} 
\renewcommand{\arraystretch}{1.2}
\begin{tabular}{@{}cccccr@{}}
\toprule
Interior anchors
& PSNR $\uparrow$
& SSIM $\uparrow$
& LPIPS $\downarrow$
& FID $\downarrow$
& FVD $\downarrow$ \\
\midrule
0   &14.38  & 0.49  &0.42 & 20.47 &110.43 \\
1  & 15.60 & 0.52 & 0.35 & 18.04 & 78.35 \\
% 2   & 17.07 & 0.56 & 0.30 & 18.53 & 65.57 \\
3 & 17.87 & 0.58 &  0.26 & 15.44 & 60.17 \\
8 & 17.85 & 0.58 & 0.26 & 17.86 & 63.44 \\
% 5   & 17.99 & 0.59 & 0.25 & {16.16} & {69.63} \\
All & {18.09} & {0.59} & {0.26} &17.37 &73.09 \\
\bottomrule
\end{tabular}
\caption{Ablation study on the number of interior anchors.}
\label{tab:anchor_number}
\end{table}

\subsection{Ablation Studies}
We ablate the spatial scene representation $\mathbf{W}^{\mathrm{sp}}$, temporal keyframe anchors $\mathbf{W}^{\mathrm{tp}}$, and motion trajectory prior $\mathbf{W}^{\mathrm{mo}}$ in Table~\ref{tab:ablation_modules}. Among individual components, $\mathbf{W}^{\mathrm{tp}}$ provides the largest baseline gain, improving PSNR from 12.96 to 16.69 and reducing FVD from 192.93 to 76.12. $\mathbf{W}^{\mathrm{mo}}$ also improves all metrics, whereas $\mathbf{W}^{\mathrm{sp}}$ primarily benefits distributional quality, reducing FID and FVD with limited changes in reconstruction metrics. Adding either $\mathbf{W}^{\mathrm{sp}}$ or $\mathbf{W}^{\mathrm{mo}}$ to $\mathbf{W}^{\mathrm{tp}}$ further improves all five metrics. The full model combines all three components and achieves the best results across all five metrics. These results indicate that temporal anchors provide the strongest target-view constraint, while the spatial and motion priors contribute complementary geometric and dynamic guidance.

We further study the effect of temporal anchor number in Table~\ref{tab:anchor_number}. Results show that increasing the number of anchors generally improves reconstruction fidelity and temporal consistency through additional egocentric view constraints. However, conditioning on all frames yields marginal improvements in PSNR, SSIM, and LPIPS but degrades the distributional and temporal metrics. We attribute this to the over-constrained nature of dense conditioning, which introduces strong appearance-level supervision, encouraging the model to rely on frame-wise correspondence rather than infer latent temporal evolution. As a result, generation becomes closer to appearance reconstruction than predictive synthesis.

\section{Conclusion}
In this work, we introduce \method{}, a structured 4D embodied world model for generating egocentric robot videos from sparse exocentric observations.
Unlike existing approaches that primarily rely on appearance-level correspondence, \method{} infers a structured representation of the embodied environment, capturing temporal state transitions, spatial scene structures, and motion evolution. Based on this representation, we construct an anchor-guided video generation framework, where synthesized egocentric keyframes, spatial representations, and motion trajectories jointly guide the generation process. Extensive experiments on real-world and simulated robotic benchmarks demonstrate that \method{} achieves improved exo-to-ego video generation quality and enhances downstream policy learning performance. These results highlight the importance of structured world representations for scalable embodied video generation and robot learning.

\FloatBarrier
\bibliographystyle{abbrvnat}
\bibliography{references}

\setcounter{secnumdepth}{2}
\setcounter{topnumber}{3}
\setcounter{totalnumber}{5}
\appendix
\section{Implementation and Training Details}
\label{app:training_details}
\method{} implementation follows the two-stage framework described in the main text. In the first stage, \method{} constructs a structured 4D embodied world representation from sparse exocentric observations by modeling temporal, spatial, and motion components. In the second stage, this representation conditions a PAI/Wan2.1-Fun-InP diffusion generator~\cite{wan2025wan} to synthesize coherent and physically consistent egocentric videos. 
Tables~\ref{tab:action_predictor_config}--\ref{tab:generator_config}
summarize the module-specific configurations on
DROID~\cite{khazatsky2025droid}. For trajectory inference, we adapt the IDM to the available exocentric views in each dataset: two synchronized views for DROID and RLBench, and one fixed view for ARX X5.

\begin{table}[ht]
\centering
\small
\renewcommand{\arraystretch}{1.12}
\setlength{\tabcolsep}{0pt}

\begin{tabularx}{0.9\linewidth}{@{}>{\bfseries\raggedright\arraybackslash}p{0.38\linewidth}@{\hspace{1.2em}}Y@{}}
\toprule
Parameter & Setting \\
\midrule
Backbone & GR00T-Dreams IDM \\
Batch size & 64 \\
Optimizer & AdamW, cosine schedule \\
Learning rate & \(1\times10^{-4}\) \\
Training steps & 150k steps \\
Chunk horizon & 16 \\
Chunk stride & 8 \\
Clip length & 81 frames \\
Hardware & 8 \(\times\) L40 GPUs \\
\bottomrule
\end{tabularx}
\caption{Trajectory predictor configuration on DROID.}
\label{tab:action_predictor_config}
\end{table}

\begin{table}[!t]
\centering
\small
\renewcommand{\arraystretch}{1.12}
\setlength{\tabcolsep}{0pt}

%---------------- Table 2 ----------------
\begin{tabularx}{0.9\linewidth}{
  @{}>{\bfseries\raggedright\arraybackslash}
  p{0.38\linewidth}@{\hspace{1.2em}}Y@{}
}
\toprule
Parameter & Setting \\
\midrule
Batch size & 64 \\
Height & 180 \\
Width & 320 \\
Optimizer & AdamW, cosine schedule \\
Grad clip & 3 \\
Learning rate & \(1\times10^{-4}\) \\
Training steps & 100k steps \\
Attention type & bidirectional cross attention \\
Source views & 2 \\
Hardware & 8 \(\times\) L40 GPUs \\
\bottomrule
\end{tabularx}

\caption{LagerNVS configuration on DROID.}
\label{tab:lagernvs_anchor_config}

\vspace{0.8em}

%---------------- Table 3 ----------------
\begin{tabularx}{0.9\linewidth}{
  @{}>{\bfseries\raggedright\arraybackslash}
  p{0.43\linewidth}@{\hspace{1.2em}}Y@{}
}
\toprule
Parameter & Setting \\
\midrule
Backbone & PAI/Wan2.1-Fun-1.3B-InP \\
Batch size & 64 \\
Clip length & 81 frames \\
Height \& Width & \(180 \times 320\) \\
Interior anchors & 3 \\
Scene tokens & 512 \\
Structured-noise weight & 0.5 \\
Trainable modules & Full \\
Optimizer & AdamW, lr = \(1\times10^{-4}\) \\
Training epochs & 10 \\
CFG & 5.0 \\
Hardware & 8 \(\times\) A800 GPUs \\
\bottomrule
\end{tabularx}

\caption{Generator configuration on DROID.}
\label{tab:generator_config}
\end{table}

\section{Video Generation Evaluation}
\label{app:baseline_details}
\subsection{Baseline Adaptation}
On DROID, we adapt the GR00T-Dreams IDM~\cite{jang2025dreamgen}
to predict an 8-D end-effector state trajectory from two synchronized
exocentric views. For the action-conditioned models,
the same IDM-inferred trajectory is used to condition both BWM and
AuroraWorld. The remaining baselines follow their respective native input interfaces,
as detailed below.

\noindent\textbf{Frame-wise synthesis.}
VGGT~\cite{wang2025vggt} and LagerNVS~\cite{szymanowicz2026lagernvs}
synthesize each target frame from the two synchronized exocentric frames
and the corresponding target egocentric camera pose provided by the
dataset. VGGT first reconstructs the 3D scene and projects it into the
target view, whereas LagerNVS directly renders the target-view image.

\begin{table*}[!t]
\setlength{\tabcolsep}{1pt}
\centering
\small
\begin{tabular}{lccccccc}
\toprule
Method & \textbf{$\mathrm{EWMScore}_{\mathrm{ego}}$} & Visual Quality & Motion Quality & Content Consistency & Physics Adherence & 3D Accuracy & Controllability \\
\midrule
WristWorld & 62.4 & 57.7 & 57.8 & 57.3 & 55.1 & 79.9 & 69.8 \\
Exo2Ego-V & 57.6 & 47.5 & 51.4 & 55.9 & 52.7 & 78.8 & 66.1 \\
BWM & 63.3 & 57.3 & 59.2 & 56.8 & 54.3 & 82.7 & \textbf{73.3} \\
SVD & 58.0 & 51.8 & 44.7 & 56.3 & 58.2 & 78.0 & 69.8 \\
LagerNVS & 53.7 & 49.3 & 39.1 & 54.2 & 44.4 & 75.7 & 63.9 \\
VGGT & 59.4 & 51.3 & \textbf{84.8} & 53.8 & 25.9 & 67.8 & 50.5 \\
\method{} & \textbf{64.6} & \textbf{59.7} & 59.0 & \textbf{58.0} & \textbf{59.7} & \textbf{83.9} & 73.1 \\
\bottomrule
\end{tabular}
\caption{WorldArena scores for DROID egocentric videos. \(\mathrm{EWMScore}_{\mathrm{ego}}\) averages the 14 applicable metrics defined in Appendix~\ref{app:worldarena_details}.}
\label{tab:worldarena}
\end{table*}

\begin{table*}[!t]
\centering
\small
\setlength{\tabcolsep}{4.8pt}
\begin{tabular}{lcccccccccccccc}
\toprule
Method & Subj. & Aes. & Img. & BG & Dyn. & Inter. & Persp. & Instr. & Sem. & Flow & Depth & Photo. & Smooth. & JEPA \\
\midrule
WristWorld & 79.5 & 29.1 & 52.8 & 90.5 & 58.7 & 55.1 & 72.6 & 53.3 & 86.3 & 32.1 & 87.2 & 1.8 & 82.5 & 91.4 \\
Exo2Ego-V & 77.7 & 27.9 & 57.4 & 89.1 & 49.7 & 52.7 & 69.0 & 45.2 & 87.0 & 24.5 & 88.6 & 0.7 & 80.0 & 57.2 \\
BWM & 79.2 & 29.4 & 49.9 & 88.4 & 59.9 & 54.3 & 73.4 & \textbf{57.7} & \textbf{88.9} & 34.6 & 91.9 & 2.8 & \textbf{82.9} & 92.6 \\
SVD & 79.0 & \textbf{31.1} & 47.7 & 88.7 & 30.1 & 58.2 & 69.2 & 53.5 & 86.1 & 31.4 & 86.7 & 1.1 & 72.8 & 76.5 \\
LagerNVS & 75.1 & 25.1 & 41.7 & 85.7 & 31.8 & 44.4 & 63.4 & 41.4 & 86.3 & 17.9 & 87.9 & 1.8 & 67.6 & 81.0 \\
VGGT & 73.6 & 29.5 & \textbf{60.7} & 87.7 & \textbf{98.1} & 25.9 & 53.7 & 21.1 & 79.9 & \textbf{100.0} & 81.9 & 0.1 & 56.3 & 63.8 \\
\method{} & \textbf{79.9} & 28.2 & 53.2 & \textbf{91.0} & 63.1 & \textbf{59.7} & \textbf{74.7} & 57.4 & 88.8 & 33.4 & \textbf{93.1} & \textbf{3.1} & 80.5 & \textbf{97.6} \\
\bottomrule
\end{tabular}
\caption{Results for the 14 retained normalized WorldArena metrics on the same DROID videos and input conditions as Table~\ref{tab:worldarena}. Subj.: subject consistency; Aes.: aesthetic quality; Img.: image quality; BG: background consistency; Dyn.: dynamic degree; Inter.: interaction quality; Persp.: perspectivity; Instr.: instruction following; Sem.: semantic alignment; Photo.: photometric consistency; Smooth.: motion smoothness.}
\label{tab:worldarena_raw}
\end{table*}

\noindent\textbf{Video generation and world modeling.}
We fine-tune SVD~\cite{blattmann2023svd} on DROID and enable
cross-attention to both exocentric videos. SVD is initialized with the
first recorded egocentric frame and receives no robot action or trajectory
condition.
For BWM~\cite{boundless2026}, we adopt its three-view input setting.
The model receives the complete sequences from both exocentric views
together with the first recorded egocentric frame, while supervision is
applied only to the target egocentric video. BWM is conditioned on the
8-D state trajectory inferred by our IDM from the source exocentric
observations, which is converted to its native action-conditioning
representation.

\noindent\textbf{Exo-to-ego video generation.}
Exo2Ego-V~\cite{liu2024exo2ego} and
WristWorld~\cite{qian2025wristworld} receive the complete synchronized
video sequences from both exocentric views without access to recorded
target egocentric frames. Following its pose-conditioned interface,
Exo2Ego-V is additionally provided with the synchronized target
egocentric camera-pose trajectory. WristWorld estimates the target-camera
pose trajectory from the source observations and uses it to project the
reconstructed 3D scene into the target egocentric view.

\subsection{WorldArena Evaluation}
\label{app:worldarena_details}

\noindent\textbf{Adapted WorldArena protocol.}
We apply the official WorldArena protocol~\cite{shang2026worldarena} using a DROID adapter, with the same metadata, instructions, evaluators, and normalization procedure for all methods. 

The protocol comprises six groups: \textbf{Visual Quality} (Image Quality, Aesthetic Quality, and JEPA), \textbf{Motion Quality} (Dynamic Degree, Flow, and Motion Smoothness), \textbf{Content Consistency} (Subject, Background, and Photometric Consistency), \textbf{Physics Adherence} (Interaction and Trajectory Accuracy), \textbf{3D Accuracy} (Depth and Perspectivity), and \textbf{Controllability} (Instruction, Semantic, and Action Following). Qwen3-VL-8B-Instruct evaluates Interaction, Perspectivity, and Instruction Following using the instruction and 16 uniformly sampled frames under deterministic decoding. All remaining metrics are computed using the released evaluators.

We exclude Action Following because it requires counterfactual action prompts and Trajectory Accuracy because it assumes globally visible motion. Neither metric is well-defined for a single egocentric view. After excluding these two metrics, we retain 14 metrics. Let
\(\mathcal{M}_{\mathrm{ego}}\) denote this set of metrics and
\(\widetilde{s}_q(m)\) denote the dataset-level normalized score of method \(m\) on metric \(q\). We compute the egocentric-view aggregate as
\begin{equation}
\mathrm{EWMScore}_{\mathrm{ego}}(m)
= \frac{1}{14}\sum_{q\in\mathcal{M}_{\mathrm{ego}}}
\widetilde{s}_q(m).
\end{equation}

Table~\ref{tab:worldarena} shows that \method{} achieves the highest 14-metric aggregate score of 64.6, exceeding BWM and WristWorld by 1.3 and 2.2, respectively. It ranks first on 7 of the 14 retained metrics and on four category-level scores: Visual Quality, Content Consistency, Physics Adherence, and 3D Accuracy. In particular, the gains in subject consistency, perspectivity, depth, and JEPA are consistent with improved preservation of scene identity and geometry. AuroraWorld does not lead Motion Quality or Controllability; VGGT obtains the highest Motion Quality score, whereas BWM performs best on Controllability. These results, together with the anchor ablations, suggest that state-conditioned anchors improve geometric and interaction consistency without uniformly improving every aspect of video dynamics.

% Table~\ref{tab:visual} reports the standard visual evaluation results on the RLBench Wrist-Withheld Split. The corresponding results on DROID are provided in Table~1 of the main paper.

% \begin{table}[t]
% \centering
% \small
% \setlength{\tabcolsep}{2.2pt}
% \begin{tabular}{lccccc}
% \toprule
% Method
% & PSNR $\uparrow$ & SSIM $\uparrow$ & LPIPS $\downarrow$ & FID $\downarrow$ & FVD $\downarrow$ \\
% \midrule
% VGGT & 10.74 & 0.33 & 0.60 & 184.46 & 1451.10 \\
% LagerNVS & 18.77 & 0.56 & 0.25 & 53.04 & 401.24 \\
% SVD & 15.88 & 0.51 & 0.45 & 70.54 & 450.37 \\
% BWM & 20.40 & \textbf{0.60} & 0.24 & 47.38 & 146.05 \\
% Exo2Ego-V & 20.37 & 0.58 & 0.19 & 31.59 & 269.39 \\
% WristWorld & 19.55 & 0.56 & 0.19 & \textbf{22.75} & 59.68 \\
% \method{} & \textbf{20.89} & 0.57 & \textbf{0.14} & 28.32 & \textbf{59.67} \\
% \bottomrule
% \end{tabular}
% \caption{Standard visual evaluation results on the RLBench Ego-Withheld Split. The arrows indicate the preferred direction for each metric.}
% \label{tab:visual}
% \end{table}

\begin{table}[t]
\centering
\small
\setlength{\tabcolsep}{2.2pt}
\begin{tabular}{lccccc}
\toprule
Method
& PSNR $\uparrow$ & SSIM $\uparrow$ & LPIPS $\downarrow$ & FID $\downarrow$ & FVD $\downarrow$ \\
\midrule
VGGT & 10.74 & 0.33 & 0.60 & 184.46 & 1451.10 \\
LagerNVS & 18.77 & 0.56 & 0.25 & 53.04 & 401.24 \\
SVD & 15.88 & 0.51 & 0.45 & 70.54 & 450.37 \\
BWM & 20.40 & \textbf{0.60} & 0.24 & 47.38 & 146.05 \\
Exo2Ego-V & 20.37 & 0.58 & 0.19 & 31.59 & 269.39 \\
WristWorld & 19.55 & 0.56 & 0.19 & \textbf{22.75} & 59.68 \\
\method{} & \textbf{20.89} & 0.57 & \textbf{0.14} & 28.32 & \textbf{59.67} \\
\bottomrule
\end{tabular}
\caption{Standard visual evaluation results on the RLBench Ego-Withheld Split.}
\label{tab:app_visual}
\end{table}

\subsection{Additional Ablations}
\label{app:additional_ablations}

\begin{figure}[t]
    \centering
    \includegraphics[width=0.9\columnwidth]{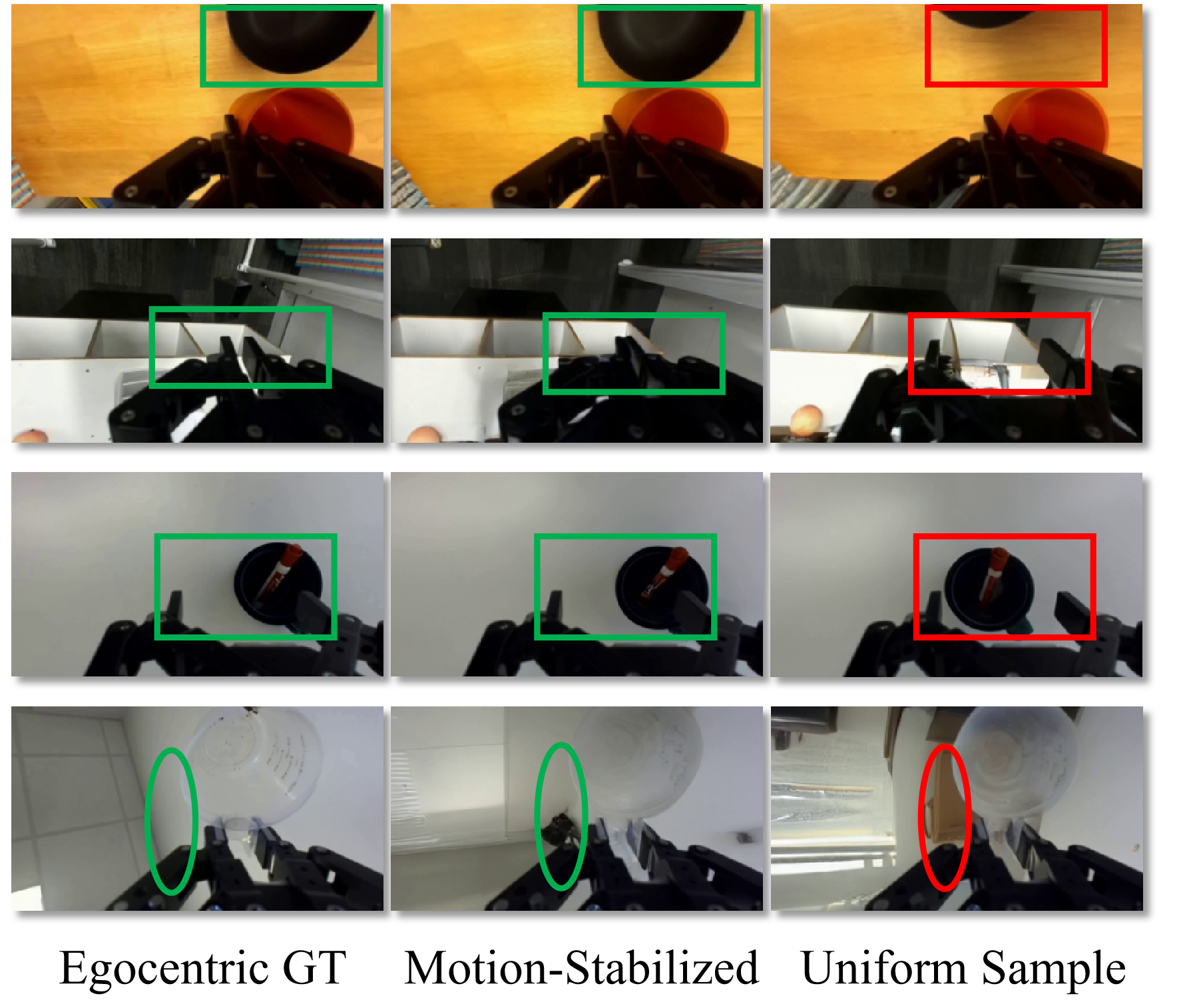}
    \caption{Qualitative comparison between uniform sampling and motion-stabilized temporal-anchor selection. Under the same five-anchor budget, motion-stabilized selection better aligns the generated sequence with task-relevant manipulation phases.}
    \label{fig:anchor_selection_qualitative}
\end{figure}

\noindent\textbf{Anchor selection.} We compare uniform sampling with motion-stabilized anchor selection under an identical budget of five anchors, including the first and last frames. Uniform sampling selects frames $\{0, 20, 40, 60, 80\}$, whereas our method retains both endpoints and selects three interior anchors near motion-transition and subsequent stabilization events. As shown in Table~\ref{tab:additional_ablations}, motion-stabilized selection consistently improves all five evaluation metrics, reducing FVD from 69.634 to 60.171. These results suggest that event-aligned anchors better capture task-relevant manipulation phases, while uniform sampling may oversample stationary intervals or miss brief motion transitions. Figure~\ref{fig:anchor_selection_qualitative} provides a qualitative comparison and shows that the selected anchors are better aligned with the target manipulation phase.

\noindent\textbf{Anchor count.}
We further study the effect of the number of temporal anchors on video quality. All sparse settings retain the first and last frames while varying the number of interior anchors. As shown in Table~\ref{tab:additional_ablations}, increasing the number of anchors substantially improves performance over endpoint-only conditioning, but the gains are not monotonic. Three interior anchors achieve the best FID and FVD, while using eight anchors or all frames provides only marginal improvements in frame-level metrics and degrades temporal quality. These results suggest that a small number of informative anchors provides a better balance between visual fidelity and temporal consistency than dense conditioning.
\begin{table}[t]
\centering
\small
\setlength{\tabcolsep}{2.0pt}
\begin{tabular}{lccccc}
\toprule
Setting & PSNR $\uparrow$ & SSIM $\uparrow$ & LPIPS $\downarrow$
& FID $\downarrow$ & FVD $\downarrow$ \\
\midrule
\multicolumn{6}{@{}l}{\textit{Anchor selection }} \\
Uniform & 17.39 & 0.572 & 0.280 & 15.754 & 69.634\\
Motion-stab. &  17.87 & 0.580 & 0.265 & 15.437 & 60.171 \\
\midrule
\multicolumn{6}{@{}l}{\textit{Number of interior anchors}} \\
0 & 14.38 & 0.492 & 0.425 & 20.466 & 110.430 \\
1 & 15.60 & 0.524 & 0.354 & 18.039 & 78.348 \\
3 & 17.87 & 0.580 & 0.265 & 15.437 & 60.171 \\
8 & 17.85 & 0.584 & 0.262 & 17.856 & 63.442 \\
All frames &18.09 & 0.585 & 0.261 & 17.369 & 73.094  \\
\bottomrule
\end{tabular}
\caption{Additional ablations on DROID. The two blocks compare temporal-anchor selection and the number of interior anchors.}
\label{tab:additional_ablations}
\end{table}

\subsection{Additional Visual Evaluation}
\label{app:rlbench_visual}
\noindent\textbf{Standard metrics on RLBench.}
We further evaluate egocentric-view generation on the RLBench Ego-Withheld split using standard reconstruction and perceptual metrics. As shown in Table~\ref{tab:app_visual}, \method{} achieves the best PSNR and LPIPS, while obtaining an FVD nearly identical to WristWorld. 

\begin{figure}[t]
\centering
\includegraphics[width=\columnwidth]{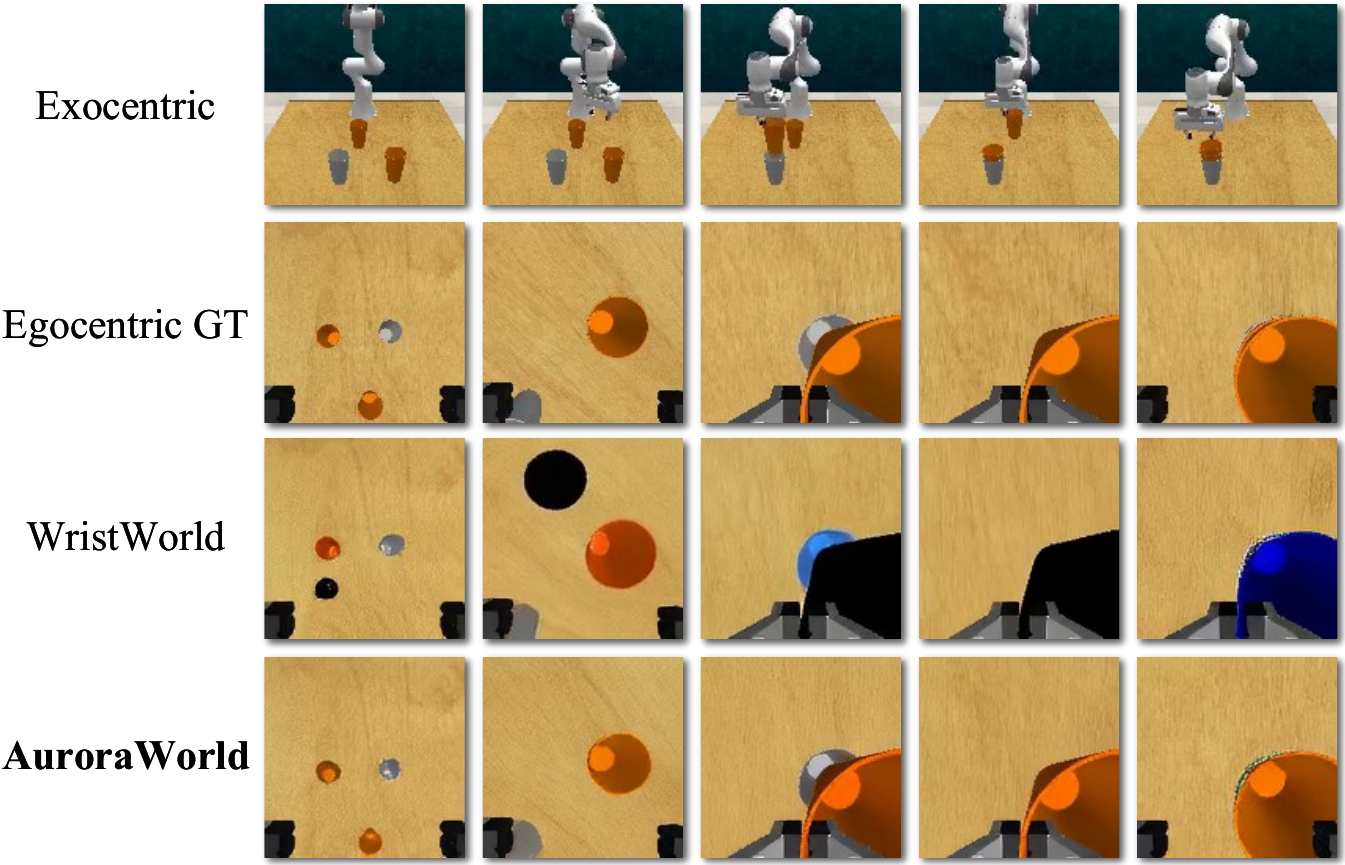}
\caption{Qualitative comparison on RLBench.}
\label{fig:rlbench_qualitative}
\end{figure}

\begin{figure*}[!t]
\centering
\includegraphics[width=1\textwidth]{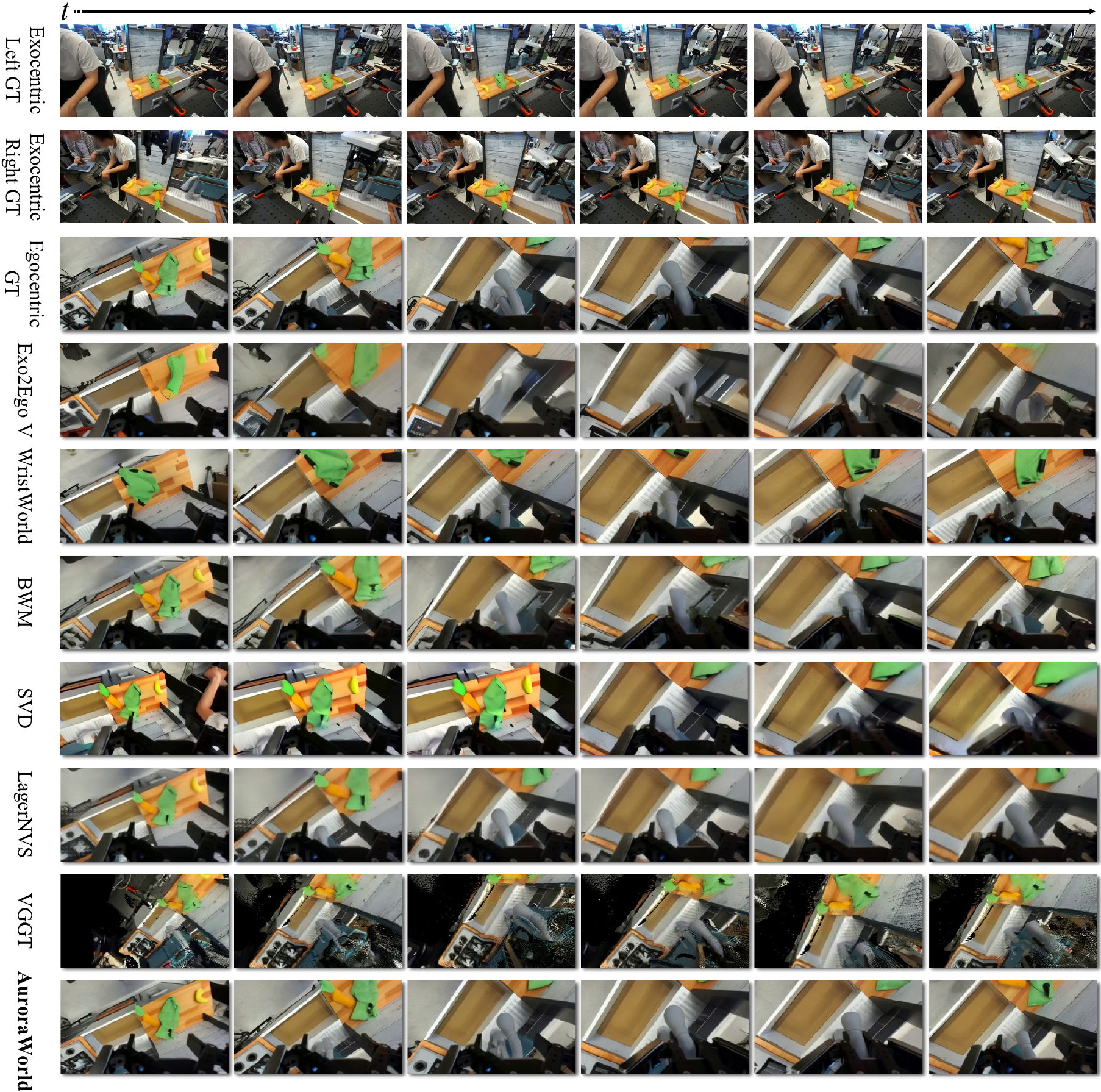}
\caption{DROID examples: exocentric left/right and egocentric-view ground truth,
Exo2Ego-V, WristWorld, BWM, SVD, LagerNVS, VGGT, and \method{}. \method{} better preserves target
geometry, appearance, and temporal layout.}
\label{fig:app_droid_qualitative}
\end{figure*}

\noindent\textbf{Qualitative comparisons.}
Figures~\ref{fig:rlbench_qualitative} and~\ref{fig:app_droid_qualitative} compare
task-relevant geometry and temporal consistency on RLBench and DROID.

\subsection{Effect of Target-Condition Source on Video Generation}
\label{app:trajectory_source_video}
By default, \method{} infers an 8-D end-effector state trajectory from synchronized exocentric observations. The inferred trajectory guides temporal-anchor selection, target-view prediction, and structured latent initialization. We evaluate how trajectory accuracy affects video generation by replacing the inferred trajectory with synchronized ground-truth states while keeping the generator weights, visual inputs, and evaluation protocol fixed. This controlled setup isolates the effect of trajectory quality on video generation.

\noindent\textbf{Within-method comparison.}
Table~\ref{tab:trajectory_source_auroraworld} first compares the two trajectory sources for \method{} on DROID, RLBench, and ARX X5. Ground-truth trajectories consistently improve PSNR, SSIM, and LPIPS across all three datasets. On DROID, replacing inferred trajectories with ground-truth trajectories improves PSNR from 17.87 to 19.17 and reduces FVD from 60.17 to 52.11. Similar improvements in frame-level reconstruction and perceptual similarity are observed on RLBench and ARX X5, demonstrating that more accurate motion trajectories provide stronger guidance for target-view synthesis. The distribution-level metrics exhibit different behavior across datasets: FID and FVD improve on DROID but increase on RLBench and ARX X5. These results indicate that improved frame-aligned reconstruction does not necessarily produce a closer match to dataset-level image and video feature distributions, which also reflect generation dynamics and dataset-specific statistics.

\begin{table}[t]
\centering
\small
\setlength{\tabcolsep}{2.0pt}
\begin{tabular}{lccccc}
\toprule
Trajectory Source
& PSNR $\uparrow$
& SSIM $\uparrow$
& LPIPS $\downarrow$
& FID $\downarrow$
& FVD $\downarrow$ \\
\midrule
\multicolumn{6}{@{}l}{\textit{DROID}} \\
Inferred
& 17.87 & 0.580 & 0.265 & 15.44 & 60.17 \\
Ground-truth
& 19.17 & 0.6218 & 0.2185 & 14.16 & 52.11 \\
\midrule
\multicolumn{6}{@{}l}{\textit{RLBench}} \\
Inferred
& 20.89 & 0.5708 & 0.1401 & 28.32 & 59.67 \\
Ground-truth
& 21.84 & 0.5845 & 0.1216 & 28.85 & 72.11 \\
\midrule
\multicolumn{6}{@{}l}{\textit{ARX X5}} \\
Inferred
& 20.01 & 0.6808 & 0.3747 & 16.91 & 149.93\\
Ground-truth
& 20.25 & 0.6827 & 0.3706 & 20.91 & 200.32 \\
\bottomrule
\end{tabular}
\caption{Video-generation metrics of \method{} under inferred and synchronized ground-truth end-effector state trajectories.}
\label{tab:trajectory_source_auroraworld}
\end{table}

\noindent\textbf{Comparison with other methods.}
We further compare the effect of target-condition quality across BWM, WristWorld, and \method{} on DROID. BWM and \method{} are conditioned on end-effector state trajectories, while WristWorld projects reconstructed 3D scenes into the egocentric image plane using frame-wise target ego-camera poses. For BWM, the state trajectory is converted to its native action-conditioning representation. In the inferred-condition setting, BWM and \method{} use the dense state trajectory inferred by our IDM from the source exocentric observations, while WristWorld uses its corresponding inferred target-view camera poses. In the ground-truth-condition setting, each method receives its synchronized ground-truth target condition. The generator weights and evaluation protocol remain fixed within each method, and BWM receives the ground-truth first egocentric frame in both settings.

Table~\ref{tab:target_condition_comparison} shows that ground-truth target conditions improve all five metrics for BWM, WristWorld, and \method{}. In particular, ground-truth conditions improve PSNR by 0.71, 1.54, and 1.30 for BWM, WristWorld, and \method{}, respectively, while reducing FVD by 11.18, 38.05, and 8.06. These consistent within-method improvements confirm that accurate target states and camera poses provide stronger motion and viewpoint guidance.

Across both target-condition settings, \method{} achieves the best results on all five metrics. With inferred conditions, \method{} obtains 17.87 PSNR and 60.17 FVD, outperforming BWM by 2.10 PSNR and 9.90 FVD points and WristWorld by 4.19 PSNR and 81.81 FVD points. With ground-truth conditions, \method{} further improves to 19.17 PSNR and 52.11 FVD, exceeding BWM by 2.69 PSNR and 6.78 FVD points and WristWorld by 3.95 PSNR and 51.82 FVD points. Moreover, AuroraWorld achieves these results without receiving a recorded egocentric frame, whereas BWM additionally receives the first recorded egocentric frame. The consistent ranking under both inferred and ground-truth conditions demonstrates the effectiveness of structured 4D modeling in translating target trajectories into geometrically accurate and temporally coherent egocentric videos.

\begin{table}[t]
\centering
\small
\setlength{\tabcolsep}{2.0pt}
\begin{tabular}{lccccc}
\toprule
Target Condition
& PSNR $\uparrow$
& SSIM $\uparrow$
& LPIPS $\downarrow$
& FID $\downarrow$
& FVD $\downarrow$ \\
\midrule
\multicolumn{6}{@{}l}{\textit{WristWorld}} \\
Inferred
& 13.68 & 0.4204 & 0.4527 & 30.27 & 141.98 \\
Ground-truth
& 15.22 & 0.5161 & 0.3981 & 25.90 & 103.93 \\
\midrule
\multicolumn{6}{@{}l}{\textit{BWM}} \\
Inferred
& 15.77 & 0.5012 & 0.3309 & 25.63 & 70.07 \\
Ground-truth
& 16.48 & 0.5319 & 0.2933 & 24.28 & 58.89 \\
\midrule
\multicolumn{6}{@{}l}{\textit{\method{}}} \\
Inferred
& \underline{17.87}
& \underline{0.5800}
& \underline{0.2648}
& \underline{15.44}
& \underline{60.17} \\
Ground-truth
& \textbf{19.17}
& \textbf{0.6218}
& \textbf{0.2185}
& \textbf{14.16}
& \textbf{52.11} \\
\bottomrule
\end{tabular}
\caption{DROID egocentric video generation under inferred and ground-truth target conditions. \textbf{Bold} marks the best result overall; \underline{underlining} marks the best result with inferred target conditions.}
\label{tab:target_condition_comparison}
\end{table}

\begin{figure}[!t]
\centering
\includegraphics[width=\columnwidth]{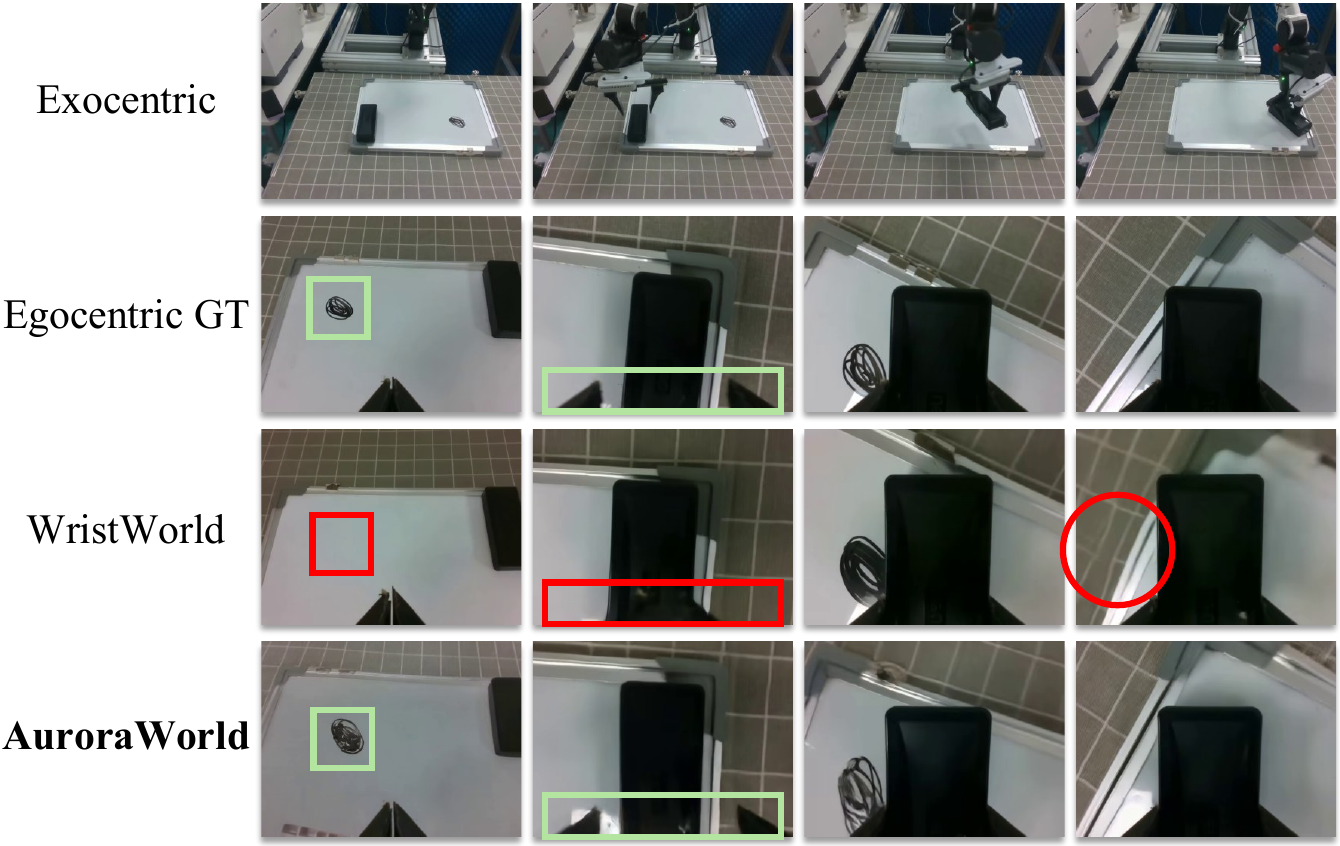}
\caption{Qualitative comparison of egocentric video generation on ARX X5.}
\label{fig:realbot_qualitative}
\end{figure}

\section{RLBench Policy Evaluation}
\label{app:rlbench_policy_details}

\subsection{Task Suite and Data Splits}
\label{app:rlbench_setup}
We use an eight-task subset of the PerAct suite~\cite{shridhar2023peract} on
RLBench~\cite{james2019rlbench}. Table~\ref{tab:rlbench_8_tasks} maps the
abbreviations in the per-task results to canonical task names.

\begin{table}[!t]
\centering
\small
\begin{tabularx}{0.9\columnwidth}{
  @{}l
  @{\hspace{0.3em}}
  >{\hsize=1.3\hsize\linewidth=\hsize\raggedright\arraybackslash}X
  @{\hspace{0.6em}}
  l
  @{\hspace{0.3em}}
  >{\hsize=0.7\hsize\linewidth=\hsize\raggedright\arraybackslash}X
  @{}
}
\toprule
Abbr. & Task & Abbr. & Task \\
\midrule
PID  & Put item in drawer          & RAD & Reach and drag \\
SBCT & Slide block to color target & TT  & Turn tap \\
PMS  & Put money in safe           & OD  & Open drawer \\
MOG  & Meat off grill              & PB  & Push buttons \\
\bottomrule
\end{tabularx}
\caption{Abbreviations and names of the eight RLBench evaluation tasks.}
\label{tab:rlbench_8_tasks}
\end{table}

\begin{figure*}[!t]
\centering
\includegraphics[width=1\textwidth]{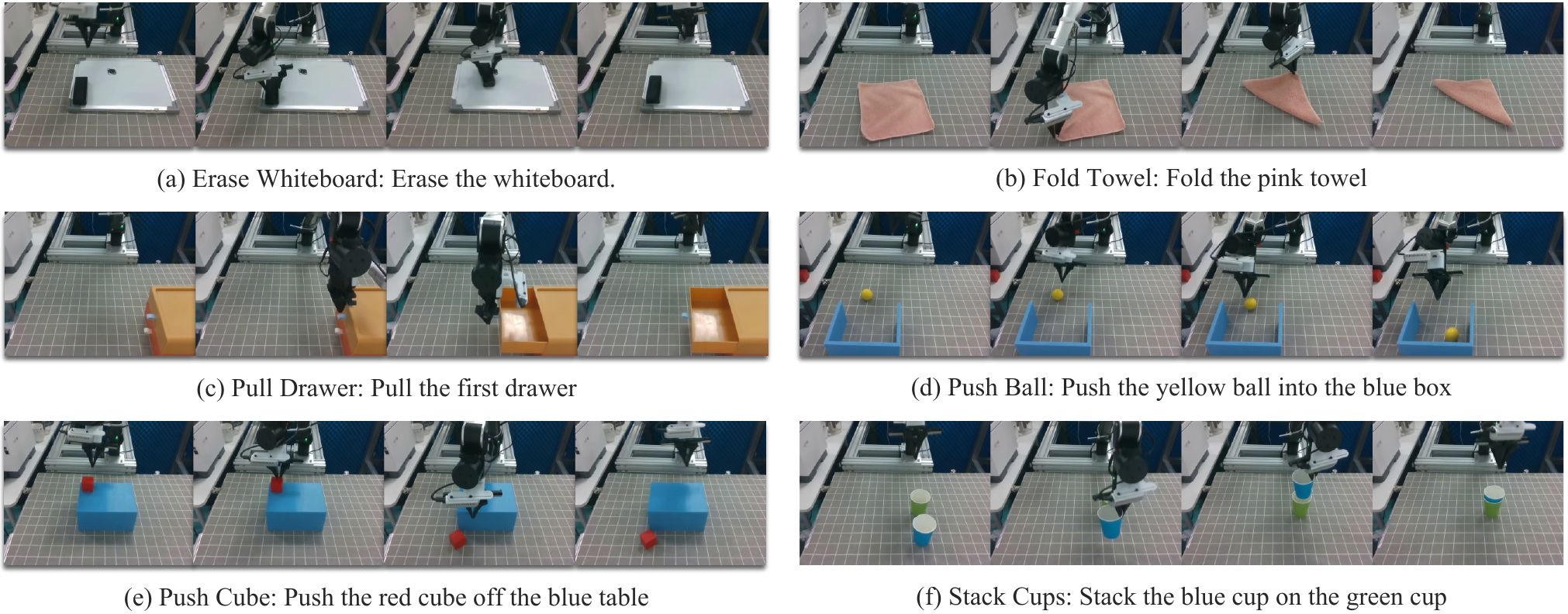}
\caption{ARX X5 performing six real-world manipulation tasks.}
\label{fig:ARX_datasets}
\end{figure*}

% For each task, we partition the 100 training demonstrations into two 50-demonstration splits, denoted Ego-Available and Ego-Withheld. We adapt the generators using only Ego-Available demonstrations with recorded egocentric observations. For Ego-Withheld demonstrations, \method{} and WristWorld synthesize egocentric observations without accessing their recorded egocentric videos. BWM additionally receives the recorded first egocentric frame of the same Ego-Withheld demonstration as a conditioning input; its remaining recorded egocentric frames are withheld. The Real Ego reference condition uses the recorded egocentric observations from Ego-Withheld demonstrations.

% In the Replacement setting, we train policies on the Ego-Withheld split using no egocentric input, WristWorld-generated views, BWM-generated views, \method{}-generated views, or recorded egocentric views. For \method{}, synchronized ground-truth end-effector trajectories guide video synthesis. In the Augmentation setting, we train policies from scratch on the union of the two splits. The Ego-Available demonstrations retain their recorded egocentric views, while the Ego-Withheld demonstrations use the same synthesized-view conditions or recorded egocentric views. We use the policy trained only on the Ego-Available split as the non-augmented baseline. Both RLBench-18 settings use recorded actions for policy supervision. The eight-task inferred-action experiment uses the Ego-Withheld Replacement split instead.
Each task has 100 training demonstrations split equally into
Ego-Available and Ego-Withheld. Generators adapt only on Ego-Available
demonstrations with recorded egocentric views. On Ego-Withheld
demonstrations, \method{} and WristWorld synthesize views without
recorded egocentric input, while BWM additionally receives the first
recorded egocentric frame of the same demonstration. The resulting
Ego-Withheld demonstrations are used for Replacement policy training.

\subsection{Policy Backbones and Training Settings}
All egocentric-source variants share this partition, architecture, initialization,
and training budget.

\noindent\textbf{RVT-2.}
We follow the official RLBench recipe for RVT-2~\cite{goyal2024rvt2}, fixing
architecture, optimization, augmentation, and policy hyperparameters.

\noindent\textbf{\(\pi_{0.5}\).}
We fine-tune the public checkpoint~\cite{black2025pi05} with OpenPI. LeRobot examples use RVT-style 9-D actions: position, quaternion, gripper, and
ignore-collisions flag. We supervise horizon-6 chunks, replan every high-level
step, and train 40k steps with global batch 64 at \(224\times224\). No-egocentric
training uses front/overhead views plus a masked zero wrist slot; other settings
retain OpenPI defaults.

\subsection{Evaluation Protocol and Results}
\noindent\textbf{Protocol.}
Generated egocentric videos are used only for offline policy training on both
RLBench and ARX X5, never at test time. During evaluation, ego-enabled
policies receive online egocentric observations---simulator-rendered
views on RLBench and live wrist-camera RGB on ARX X5---whereas
w/o Ego uses exocentric observations only.

\noindent\textbf{Per-task results.}
% Each checkpoint runs 10 repeats on the same 25 test episodes per task, with at
% most 25 steps. Binary success yields 4,500 rollouts per setting and
% 250 per task. Table~\ref{tab:rlbench_18_full} gives per-task results for the retained baselines and recorded-egocentric references. The GT-trajectory \method{} result under Replacement is reported in Table~\ref{tab:rlbench_policy}; the eight-task results using inferred action targets are reported in Table~\ref{tab:rlbench_predicted_actions}.
% Each checkpoint is evaluated over 10 repeats of the same 25 test episodes per task, with at most 25 steps per rollout. This gives 250 rollouts per task, or 4,500 per setting on RLBench-18 and 2,000 per setting on the eight-task subset. Table~\ref{tab:rlbench_18_full} reports per-task results for both policy backbones under Replacement and Augmentation, including \method{} with ground-truth trajectories for video synthesis. Table~\ref{tab:app_rlbench_policy} summarizes these results; its Replacement averages also appear in Table~\ref{tab:rlbench_policy}. The eight-task results with inferred action supervision are reported in Table~\ref{tab:rlbench_predicted_actions}.
Each checkpoint is evaluated over 10 repeats of the same 25 test episodes per task, with at most 25 steps per rollout. This gives 250 rollouts per task, or 2,000 rollouts per setting on RLBench. Table~\ref{tab:rlbench_predicted_actions_full} details the RLBench results summarized in Table~\ref{tab:arx_real_robot}, with task abbreviations defined in Table~\ref{tab:rlbench_8_tasks}.

\begin{table}[H]
\centering
\small
\setlength{\tabcolsep}{2pt}
\renewcommand{\arraystretch}{1.05}
\begin{tabular*}{\columnwidth}{@{\extracolsep{\fill}}l*{6}{r}@{}}
\toprule
& \multicolumn{3}{c}{RVT-2} & \multicolumn{3}{c}{\(\pi_{0.5}\)} \\
\cmidrule(lr){2-4}\cmidrule(lr){5-7}
Task & WW & BWM & Ours & WW & BWM & Ours \\
\midrule
PID  & 69.2 & 79.2 & 73.6 & 5.2  & 8.8  & 15.6 \\
RAD  & 69.6 & 55.6 & 68.0 & 77.6 & 63.2 & 65.6 \\
TT   & 56.8 & 67.6 & 76.8 & 24.8 & 27.2 & 28.0 \\
SBCT & 33.6 & 37.2 & 49.6 & 75.6 & 65.2 & 70.4 \\
OD   & 50.4 & 47.6 & 55.6 & 13.6 & 10.8 & 17.2 \\
PMS  & 40.0 & 43.6 & 52.0 & 28.8 & 43.6 & 38.0 \\
PB   & 69.6 & 86.4 & 82.0 & 31.6 & 41.2 & 36.0 \\
MOG  & 55.6 & 71.6 & 54.4 & 37.6 & 41.6 & 50.4 \\
\midrule
Avg. & 55.60 & 61.10 & \textbf{64.00} & 36.85 & 37.70 & \textbf{40.15} \\
\bottomrule
\end{tabular*}
\caption{Per-task RLBench policy success (\%) for the eight-task inferred-action Replacement experiment (250 rollouts per task). WW: WristWorld; BWM: Boundless World Model.}
\label{tab:rlbench_predicted_actions_full}
\end{table}

\section{Real-Robot ARX X5 Evaluation}
\label{app:real_robot_details}

\begin{table}[t]
\centering
\small
\setlength{\tabcolsep}{1.2pt}
\renewcommand{\arraystretch}{1.2}
\begin{tabular}{lccccc}
\toprule
Task & Avail. only & +WW & +BWM & +Ours & Real Ego \\
\midrule
Fold Towel       & 16/25 & 17/25 & 23/25 & 21/25 & 22/25 \\
Pull Drawer      & 14/25 & 15/25 & 16/25 & 20/25 & 20/25 \\
Stack Cups       & 13/25 &  6/25 & 10/25 &  8/25 & 16/25 \\
Push Ball & 14/25 & 17/25 & 17/25 & 18/25 & 20/25 \\
Push Cube   & 16/25 & 22/25 & 19/25 & 22/25 & 23/25 \\
Erase Whiteboard & 10/25 & 10/25 & 13/25 & 17/25 & 16/25 \\
\midrule
Average & 55.3\% & 58.0\% & 65.3\% & 70.7\% & 78.0\% \\
\bottomrule
\end{tabular}
\caption{ARX X5 policy success under data augmentation. Each entry reports
successful trials out of 25 physical trials. WW denotes WristWorld.}
\label{tab:arx_augmentation}
\end{table}

\subsection{Real-World Setup and Data Collection}
\noindent\textbf{Hardware setup.}
Our real-world experiments use the ARX X5 single-arm robot equipped with two
Intel RealSense D435i cameras. One camera provides a fixed exocentric view, while the other is wrist-mounted to capture egocentric observations. The IDM uses this fixed exocentric view to infer end-effector trajectories.

\noindent\textbf{Data collection.}
We collect expert demonstrations using leader--follower teleoperation.
Both cameras record synchronized RGB observations at \(640\times480\) and
15\,Hz, together with robot states. We collect 100 demonstrations
for each of six tasks, yielding 600 episodes. For each task, the demonstrations
are divided into disjoint 50-episode Ego-Available and Ego-Withheld splits.
The former adapts the video generator, which synthesizes egocentric observations for
the latter without accessing its recorded egocentric videos.

\noindent\textbf{Real-world tasks.}
The six tasks cover deformable-object manipulation, articulated-object
interaction, stacking, pushing, and surface contact: \textit{Fold towel},
\textit{Pull drawer}, \textit{Stack cups}, \textit{Push yellow ball},
\textit{Push cube}, and \textit{Erase whiteboard}.
Figure~\ref{fig:ARX_datasets} shows representative trajectories.

\subsection{Policy and Evaluation Protocol}
\label{app:arx_policy_protocol}
We evaluate MOTIF~\cite{zhi2026motiflearningactionmotifs} on all six tasks using exocentric RGB, robot
state, language, and optionally wrist RGB. Architecture and training remain
fixed across conditions. All ARX X5 policies use training action targets derived from IDM-inferred end-effector trajectories. Each condition receives 25 physical trials per task
under identical initialization, termination, and success criteria.

\subsection{Additional Evaluation Results}
\noindent\textbf{Data augmentation.} Beyond the egocentric-replacement experiment, we augment the 300
Ego-Available demonstrations with the 300 Ego-Withheld demonstrations.
Table~\ref{tab:arx_augmentation} compares WristWorld, BWM, and \method{} with
Ego-Available-only and Real-Ego training. Evaluation follows Section~D.2.

\noindent\textbf{Qualitative video comparison.} Figure~\ref{fig:realbot_qualitative} compares generated egocentric views on
ARX X5.

% \subsection{Qualitative Results}
% Figure~\ref{fig:realbot_qualitative} compares generated egocentric views on
% ARX X5.

\section{Limitations}
\label{app:limitations}

\method{} relies on several upstream predictions whose errors may propagate through the generation pipeline. In particular, inaccuracies in the IDM trajectory can affect temporal-anchor selection, target-camera prediction, and structured-prior construction, as indicated by the trajectory-source comparison. Sparse exocentric observations may also be insufficient to resolve the target view under severe occlusion, rapid contact dynamics, or tasks requiring fine-grained spatial alignment, causing rendered-anchor errors to introduce incorrect geometry or temporal phase information into the generator. Consequently, policies trained with synthetic egocentric observations do not consistently match the performance of those trained with recorded egocentric views.
Although \method{} improves average policy performance in both simulation and
real-robot experiments, noticeable gaps remain on tasks involving precise
contact, occlusion, or multi-object spatial reasoning.

\end{document}